\documentclass[10pt,journal,compsoc]{IEEEtran}
\ifCLASSOPTIONcompsoc
  \usepackage[nocompress]{cite}
\else
  \usepackage{cite}
\fi
\ifCLASSINFOpdf
\else
\fi
\usepackage[utf8]{inputenc} 
\usepackage[T1]{fontenc}    
\usepackage{hyperref}       
\usepackage{url}            
\usepackage{booktabs}       
\usepackage{amsfonts}       
\usepackage{nicefrac}       
\usepackage{microtype}      
\usepackage{amsmath,amssymb,amsfonts}
\usepackage{algorithmic}
\usepackage{graphicx}
\usepackage{textcomp}
\usepackage{xcolor}
\usepackage{amsthm}
\usepackage{enumerate}
\usepackage{multirow}
\usepackage{caption}
\usepackage{subfigure}
\usepackage{balance}
\usepackage{bbm}
\usepackage{array}
\usepackage{tabularx}
\usepackage{tabu}
\usepackage{wrapfig}
\usepackage{cite}

\begin{document}
	
\title{Act Like a Radiologist: Towards Reliable Multi-view Correspondence Reasoning for Mammogram Mass Detection}

\author{
	Yuhang Liu$^*$, Fandong Zhang$^*$, Chaoqi Chen, Siwen Wang, Yizhou Wang and Yizhou Yu,~\IEEEmembership{Fellow,~IEEE}
	\IEEEcompsocitemizethanks{
		\IEEEcompsocthanksitem 
		Corresponding author: Yizhou Yu. * indicates equal contribution.\\
		\IEEEcompsocthanksitem 
		Y. Liu, C. Chen, and S. Wang are with the AI Lab, Deepwise Healthcare,
		Beijing 100080, China.\\
		\IEEEcompsocthanksitem F. Zhang is with the Center for Data Science, Peking University, Beijing 100871, China.\\ 
		\IEEEcompsocthanksitem Y. Wang is with the Center on Frontiers of Computing Studies, Dept. of Computer Science \& Technology, Advanced Institute of Information Technology, Peking University, Beijing 100871, China.\\ 
		\IEEEcompsocthanksitem  Y. Yu is with Department of Computer Science, The University of Hong Kong, Pokfulam, Hong Kong. E-mail: yizhouy@acm.org}}

\markboth{IEEE TRANSACTIONS ON PATTERN ANALYSIS AND MACHINE INTELLIGENCE}%
{Shell \MakeLowercase{\textit{et al.}}: Bare Demo of IEEEtran.cls for Computer Society Journals}
%


\IEEEtitleabstractindextext{
	\begin{abstract}
		Mammogram mass detection is crucial for diagnosing and preventing the breast cancers in clinical practice.
		The complementary effect of multi-view mammogram images provides valuable information about the breast anatomical prior structure and is of great significance in digital mammography interpretation. 
		However, unlike radiologists who can utilize the natural reasoning ability to identify masses based on multiple mammographic views, how to endow the existing object detection models with the capability of multi-view reasoning is vital for decision-making in clinical diagnosis but remains the boundary to explore.
		In this paper, we propose an Anatomy-aware Graph convolutional Network (AGN), which is tailored for mammogram mass detection and endows existing detection methods with multi-view reasoning ability.
		The proposed AGN consists of three steps. 
		Firstly, we introduce a Bipartite Graph convolutional Network (BGN) to model the intrinsic geometric and semantic relations of ipsilateral views. 
		Secondly, considering that the visual asymmetry of bilateral views is widely adopted in clinical practice to assist the diagnosis of breast lesions, 
		we propose an Inception Graph convolutional Network (IGN) to model the structural similarities of bilateral views. 
		Finally, based on the constructed graphs, the multi-view information is propagated through nodes methodically, which equips the features learned from the examined view with multi-view reasoning ability. 
		Experiments on two standard benchmarks reveal that AGN significantly exceeds the state-of-the-art performance.
		Visualization results show that AGN provides interpretable visual cues for clinical diagnosis.
	\end{abstract}
	
	}

\maketitle

\IEEEdisplaynontitleabstractindextext

%
\IEEEpeerreviewmaketitle

\IEEEraisesectionheading{\section{Introduction}\label{sec:introduction}}

%
%
%
%

\begin{figure*}[t]
	\centering
	\subfigure[Examined view]{\label{fig:1}\includegraphics[height=0.32\linewidth]{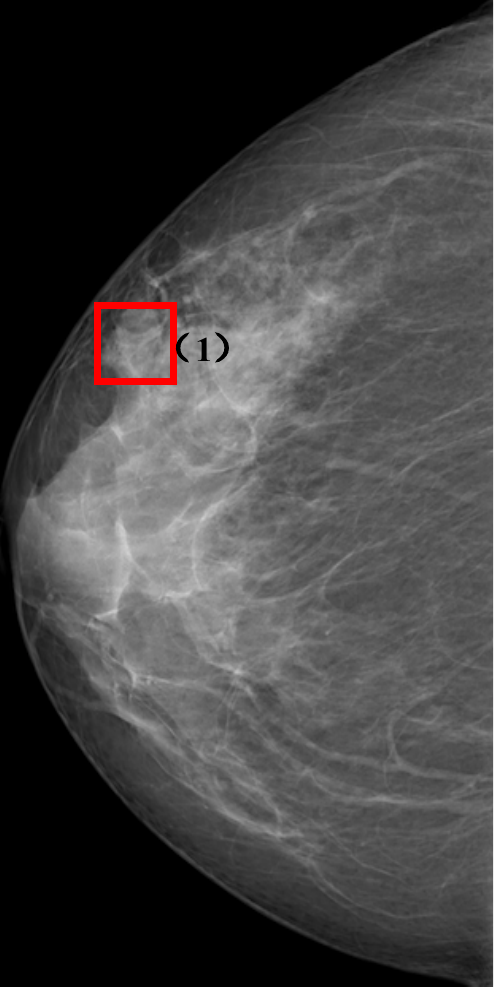}}
	\subfigure[Contralateral view ]{\label{fig:2}\includegraphics[height=0.32\linewidth]{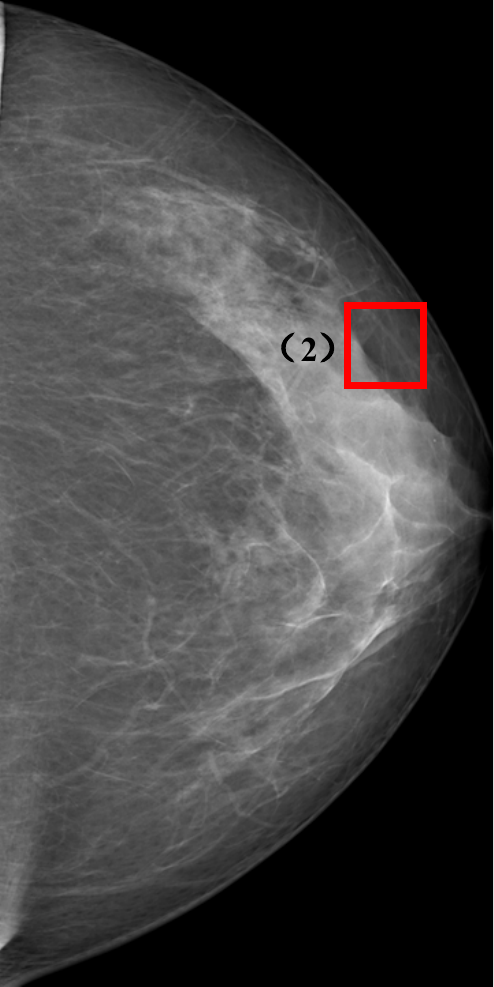}}
	\subfigure[Auxiliary view]{\label{fig:6}\includegraphics[height=0.32\linewidth]{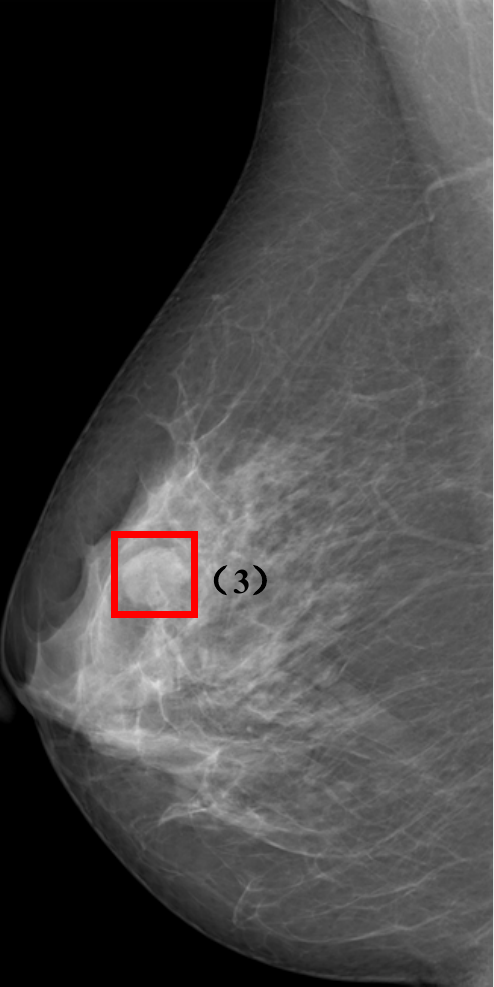}}
	\subfigure[Zoom-in patches]{\label{fig:6}\includegraphics[height=0.32\linewidth]{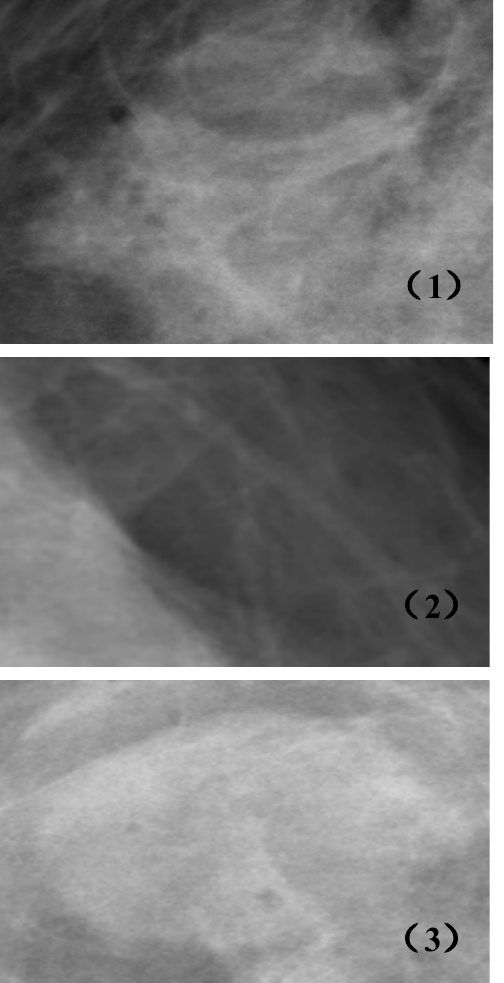}}
	\subfigure[Ideal projection model]{\label{fig:7}\includegraphics[height=0.32\linewidth]{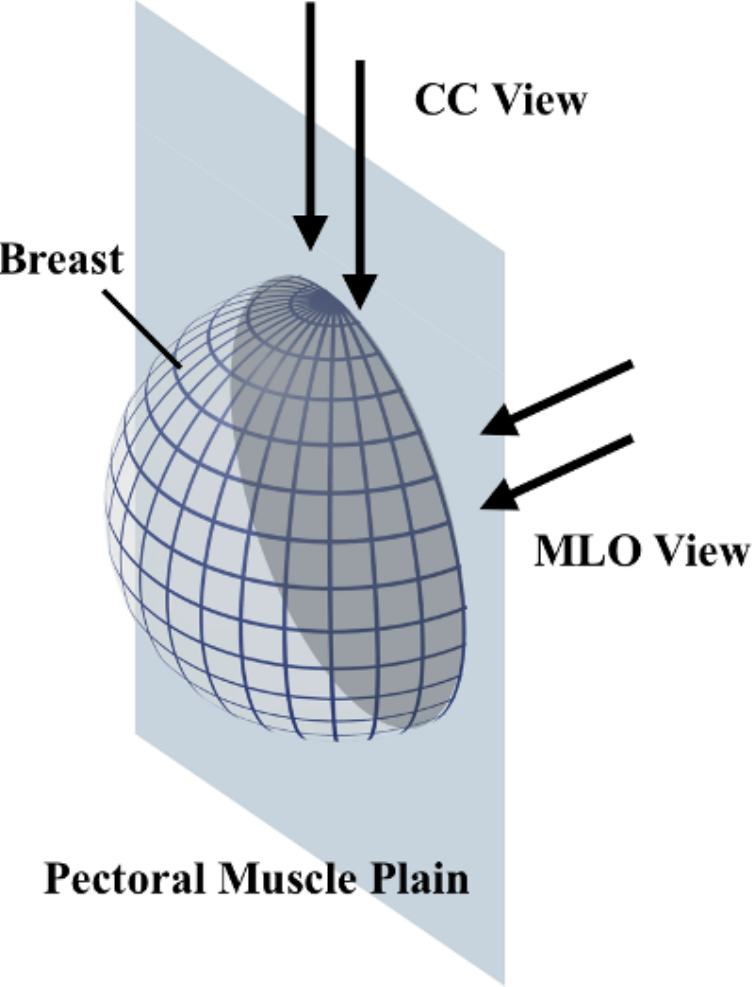}}
	
	\caption{An illustration of the relation among mammography views. Standard mammography screening takes a CC view and an MLO view for each breast. Figure (a)-(c) represent the examined view (i.e., CC view of the right breast), the contralateral view (i.e., CC view of the left breast) and the auxiliary view (i.e., MLO view of the right breast) of a specific instance. The examined view represents the view where the detection is performed. Figure (a) and (b) form a bilateral pair, and are roughly symmetric since they have similar gland background, breast shape and breast size. Figure (a) and (c) form an ipsilateral pair, and provide complementary information to represent the 3D anatomical structure. Patch (1) and (3) in the corresponding images refer to a mass lesion instance, while Patch (2) locates similarly as Patch (1) in the contralateral image. Figure (d) offers zoom-in versions of these patches. Figure (e) stands for an ideal projection model of mammography. We can see that the CC view is a top-down view of the breast, while the MLO view is a side view taken at a certain angle along the pectoral muscle plane.}
	\label{fig:motivation}
\end{figure*}

\IEEEPARstart{B}{reast} cancer, which has the highest incidence and mortality rates among women~\cite{siegel2014cancer}, is one of the leading cause of cancer deaths worldwide. 
Screening mammography has demonstrated strong efficacy in reducing the breast cancer mortality especially at the early stage~\cite{sickles1997breast}. 
The detection of masses based on mammograms is a key step for diagnosing breast cancer in clinical practice.
Nevertheless, masses can be partially obscured by high-intensity compacted glands especially in dense breasts, which imposes great challenges on radiologists and computer-aided detection (CAD) systems for the detection of mass from mammography.
To better assist clinical diagnosis, mammogram mass detection is typically based on multiple views on both breasts. 
Specifically, as shown in Figure~\ref{fig:motivation}, a cranio-caudal (CC) view (i.e., a top-down view of the breast) and a mediolateral oblique (MLO) view (i.e., a side view of the breast taken at a certain angle) are taken for both breasts. 
Comparing ipsilateral views (i.e. both CC and MLO views of the same breast)  helps to analyze 3D structure of masses. Besides, since bilateral views (i.e., same view of both breasts) usually share a similar breast structure, asymmetric regions of bilateral views  are more likely to be masses   (Detailed definitions with respect to mammogram views will be described in Section~\ref{sec:preliminaries}).
Therefore, the complementary effect of multi-view mammogram images is capable of providing valuable information regarding the breast anatomical prior structure, which is of great significance in digital mammography interpretation. 

In terms of exploiting the relations of multi-view images, prior works can be roughly divided into two categories: ipsilateral view based and bilateral view based methods. 
For the ipsilateral view based modeling, 
an intuitive approach is to leverage the relation networks~\cite{hu2018relation, WangNon} to model the ipsilateral inter-image non-local relations. 
For example, 
CVR-RCNN~\cite{ma2019cross} cascades a relation module \cite{hu2018relation} to the second stage of Faster RCNN~\cite{ren2015faster} to model the inter-proposal relations between CC and MLO views.
However, compared to radiologists who can assist reasoning with domain knowledge,  relation learning lacks clear constraints, i.e., the ipsilateral geometric and semantic relations are not explicitly taken into consideration. Thus, the learned relations may be incapable of precisely modeling the ipsilateral relations. 
In addition, it should be noted that such relation module relies, to a large extent, on the quality of region proposals at the first stage. When there exists the situation of severe gland occlusions, the performance will drop significantly. For the bilateral views~\cite{chen2020anatomy}, a recent work, i.e., CBN~\cite{liu2019unilateral}, proposes to fuse features of bilateral views with added tolerance of geometric distortions. However, like CVR-RCNN, CBN is based on RPN proposals~\cite{ren2015faster}, which also suffers from the proposal-missing problem.

During the exploration, we notice that radiologists can explicitly utilize the natural reasoning ability to identify masses by observing different mammographic views. The diagnosis flow adopted by the radiologists consists of the following three steps.  
(1) Determine the suspicious regions based on the examined view. 
(2) Search for compatible regions in the auxiliary view on the basis of appearances and locations, and contrast the corresponding regions based on the bilateral pair (i.e. examined view and contralateral view). 
(3) Make a diagnosis with respect to the suspicious regions according to two visual observations: (i) reasonable correspondences are found in the auxiliary view; and (ii) the regions in the bilateral view are symmetric.
While the multi-view region-based reasoning procedure plays a key role for mammogram mass detection, most existing methods~\cite{lin2017focal,wu2019deep,liu2019unilateral,kooi2017large,DhungelA2017Mammogrammass,diniz2018detection,ribli2018detecting,jung2018detection,cao2019deeplima} focus on improving the detection accuracy in a single view. How to endow the existing object detection models with the capability of multi-view reasoning is vital for decision-making in clinical diagnosis but remains the boundary to explore.

Inspired by the aforementioned discussions, in this paper, we delve into the multi-view reasoning problem and propose an Anatomy-aware Graph convolutional Network (AGN), which is tailored for mammogram mass detection and endows detection models with multi-view reasoning ability. 
By jointly reasoning the correspondence relations among multiple mammogram views, AGN learns the intact multi-view information in an end-to-end manner during training. 
The input of AGN is the extracted features of multi-view images from the backbone network, and the output is the enhanced features of the examined view.
As a general method, AGN can be easily plugged into any modern object detection frameworks~\cite{ren2015faster,he2017mask,yang2019reppoints} without modifying their original network architectures. 

To be specific, the proposed AGN is comprised of the following three steps.
\textbf{Firstly}, we introduce a novel Bipartite Graph convolutional Network (BGN) to model the \textit{intrinsic geometric and semantic relations}  of ipsilateral views. 
The construction of bipartite graph nodes aims at modeling the region-level correspondences between views. Each node represents a region with relatively consistent locations across breast instance. 
The bipartite graph edges are constructed to characterize the relation between nodes across views in two aspects: geometric constraints and appearance similarities. \textbf{Secondly}, considering that the visual asymmetry of bilateral views is widely adopted in clinical practice to assist the diagnosis of breast lesions~\cite{chen2020anatomy}, 
we propose an Inception Graph convolutional Network (IGN) to model the \textit{structural similarities of bilateral views}. 
IGN targets on contrasting the bilateral mammogram views based on the assumption that asymmetric regions are more likely to be masses. Technically, it forms multi-branch graph connections between each node and its nearest neighbors, which strengthens the robustness of learned representations against inherent geometric distortions and forms an Inception-like structure~\cite{szegedy2015going}.
\textbf{Finally}, based on the constructed bipartite and inception graphs, the multi-view information is propagated through nodes methodically after several layers of graph convolutions, which equips the features from the examined view with multi-view reasoning ability. 

Compared to prior works that leverage weak or even no reasoning constraints,
our AGN explicitly learns a customized multi-view reasoning model from both ipsilateral and bilateral views.
In addition, the proposed BGN and IGN enhance the backbone features before the region proposal step, which helps to mitigate the proposal-missing problem.
We evaluate the effectiveness of the proposed AGN on two mammogram mass detection benchmarks, i.e., a public dataset (DDSM~\cite{heath2000digital}) and a multi-center in-house dataset. Our experiments reveal that the proposed algorithm significantly exceeds state-of-the-art performance on benchmark datasets. Moreover, visualization results demonstrate that the proposed AGN provides reasonable and interpretable visual cues for clinical diagnosis. 

The contributions of our work are summarized as follows: 
Firstly, to the best of our knowledge, this is the first work that explicitly exploits the multi-view graphical correspondence for mammogram mass detection.
Secondly, we propose a bipartite graph convolutional network, which is capable of performing reasoning about ipsilateral correspondences and modeling both geometric constraints and visual similarities between ipsilateral views. 
Lastly, we design an inception graph convolutional network, which models the structural similarities between bilateral views and enhances the robustness of learned representations relying on a \textit{priori} that asymmetric regions are more likely to be masses.

This paper substantially extends our conference paper~\cite{Liu_2020_CVPR} from four major aspects.
\textbf{(1)} Besides utilizing ipsilateral views~\cite{Liu_2020_CVPR}, our AGN further considers the complementary effect of bilateral views to learn intact multi-view information of mammogram, which helps to make more comprehensive and precise clinical decisions. Specifically, we propose a novel inception graph convolutional network for modeling the structural similarities of bilateral views. 
\textbf{(2)} We enhance the mechanism of correspondence reasoning to fit the multi-view modeling scenario. 
\textbf{(3)} We conduct more experiments and ablation studies with respect to the enhanced network architecture, and include updated experimental results on a larger in-house multi-center dataset.
\textbf{(4)} We provide more complete introduction and analysis for the proposed multi-view correspondence reasoning network, as well as more elaborated implementation details.

The rest of the paper is organized as follows:
Section~\ref{sec:related_work} gives a brief review of related works. Section~\ref{sec:preliminaries} illustrates the preliminaries of mammogram views. Section~\ref{sec:method} demonstrates the details of the proposed AGN. Experimental results and feature visualization are described in Section~\ref{sec: experiment}.  Section~\ref{sec:conclusion} draws the conclusion.

\section{Related Work}
\label{sec:related_work}
\subsection{Mammogram Mass Detection}
Existing works on mammogram mass detection \cite{cheng2006approaches} can be coarsely classified into two categories: traditional and deep-learning-based approaches.
Traditional approaches rely on the handcrafted features to identify masses from mammogram images~\cite{mudigonda2001detection, tai2014automatic}. 
The pipeline of these approaches usually includes two stages. 
The first stage, which aims at extracting region proposals to recall most masses, is to generate candidates. Based on the assumption that mass regions are brighter than background, region proposals are obtained using thresholding, clustering, bilateral image subtraction, etc~\cite{brzakovic1990approach, mendez1998computer, li2001computerized, zhen2001artificial}. To further enlarge the intensity difference between mass regions and background, pre-processing methods are introduced, such as histogram equalization, exponent functions, etc. 
The second stage is to reduce the false positives. Numerous approaches~\cite{wei2005computer, tai2014automatic} resort to the handcrafted patterns to represent mass boundaries, textures or shapes. 
The traditional approaches suffer from the following limitations.
Firstly, traditional approaches, which are based on handcrafted features, have weak representation ability and cannot be end-to-end trainable. 
Secondly, the generated candidates are very likely to contain many false positives, increasing the difficulty of optimizing the classifier in the second stage. 
Lastly, the second stage does not contain the localization step. Thus, the predicted locations of masses may largely deviate from their bounding boxes.  

In the past decade, the renaissance in deep learning has greatly promoted the development of medical image computing~\cite{xu2018less, guo2018deep, dhungel2015automated, zhang2019cascaded,  wang2017central, yellin2018multi, liu2019align, wu2019learning, li2018thoracic, shen2017multi}.
Mass detection has achieved remarkable success by virtue of deep convolutional neural networks~\cite{lin2017focal,wu2019deep, liu2019unilateral}. 
A typical solution is to apply deep convolutional networks for reducing false positives~\cite{kooi2017large, DhungelA2017Mammogrammass, diniz2018detection}. However, these models cannot be end-to-end trainable, and thus result in inferior performance. 
To tackle this issue, researchers attempt to leverage the off-the-shelf modern object detectors, such as Faster R-CNN \cite{ren2015faster}, FPN \cite{lin2017feature}, Mask R-CNN \cite{he2017mask}, for the mammogram mass detection~\cite{ribli2018detecting, jung2018detection, cao2019deeplima}. 
Despite their general efficacy, the complementary of multi-view mammogram images are not taken into considered. 
Ma \textit{et al}.~\cite{ma2019cross} propose to model the ipsilateral property and introduce a relation module \cite{hu2018relation} to the Faster RCNN \cite{ren2015faster}, aiming to learn ipsilateral inter-proposal relations. 
However, the relation learning lacks clear constraints, i.e., the ipsilateral geometric and semantic relations are not explicitly considered. Thus, the learned relations may be incapable of precisely modeling the between-image correlations. In addition, such relation module heavily relies on the quality of region proposals at the first stage. When encountering the situation of severe gland occlusions, the detection performance will drop significantly.
By explicitly leveraging the domain knowledge of specific image modalities, 
AGN has powerful multi-view reasoning ability, which significantly improves the localization ability of backbone features. MommiNet~\cite{yang2020momminet} is a concurrent work that proposes to simultaneously perform end-to-end bilateral and ipsilateral analysis of mammogram images. However, they do not consider the graphical correspondence among different mammographic views, which is important for the success of multi-view reasoning.

\subsection{Visual Reasoning based on Graph Convolutional Network}
Visual reasoning attempts to merge distinct information (interactions) among objects (scenes), and has been extensively explored in computer vision problems, 
e.g., image classification \cite{almazan2014word, marino2016more}, object detection \cite{chen2018iterative, jiang2018hybrid}, semantic segmentation \cite{li2018beyond} and other visual understanding tasks~\cite{lampert2009learning,misra2017red, frome2013devise, mao2015learning, redmon2016you}. 
The typical paradigm of visual reasoning is to incorporate the object relations or attributes into different vision tasks~\cite{akata2013label, almazan2014word, hu2018relation}. 
For example, Akata \textit{et al}.~\cite{akata2013label} solve the attribute-based image classification problem by regarding it as a problem of reasoning in the attribute-embedded space.

Recently, Graph Convolutional Network (GCN)~\cite{zhou2018graph} has been introduced for visual reasoning tasks due to its representation power for non-Euclidean data and reasoning power from domain knowledge.
Li \textit{et al}.~\cite{li2018beyond} propose a set of graph convolutional units to learning graph representations from 2D visual data.  
However, the information propagation during the representation learning process will inevitably introduce noisy signals since all semantic correspondences are jointly considered.  
Moreover, the reasoning procedure is uncontrolled and implicit, which impairs the performance and the interpretability.  
Gao \textit{et al}.~\cite{gao2019graph} strengthen the expressive power of learned features for the visual tracking task by introducing a spatial-temporal GCN. 
However, the construction of graph nodes is based on the uniform grids, which are very sensitive to the variations of object scales, image shapes, geometric structures, etc.
Considering the semantic dependencies among different objects, 
Xu \textit{et al}.~\cite{xu2019reasoning} propose to exploit the human commonsense knowledge to reason with a class-to-class prior. 
However, the constructed knowledge graph remains relatively fixed, and thus may fail to adapt to the more complex scenarios. 
In addition, it is incapable of extending to the single-category detection problems (e.g., mass detection).
The reasoning procedure that radiologists read mammograms provides more explicit guidance, which motivates us to design a more customized algorithm with domain knowledge.

\begin{figure*}
	\begin{center}
		\includegraphics[height=0.68\linewidth]{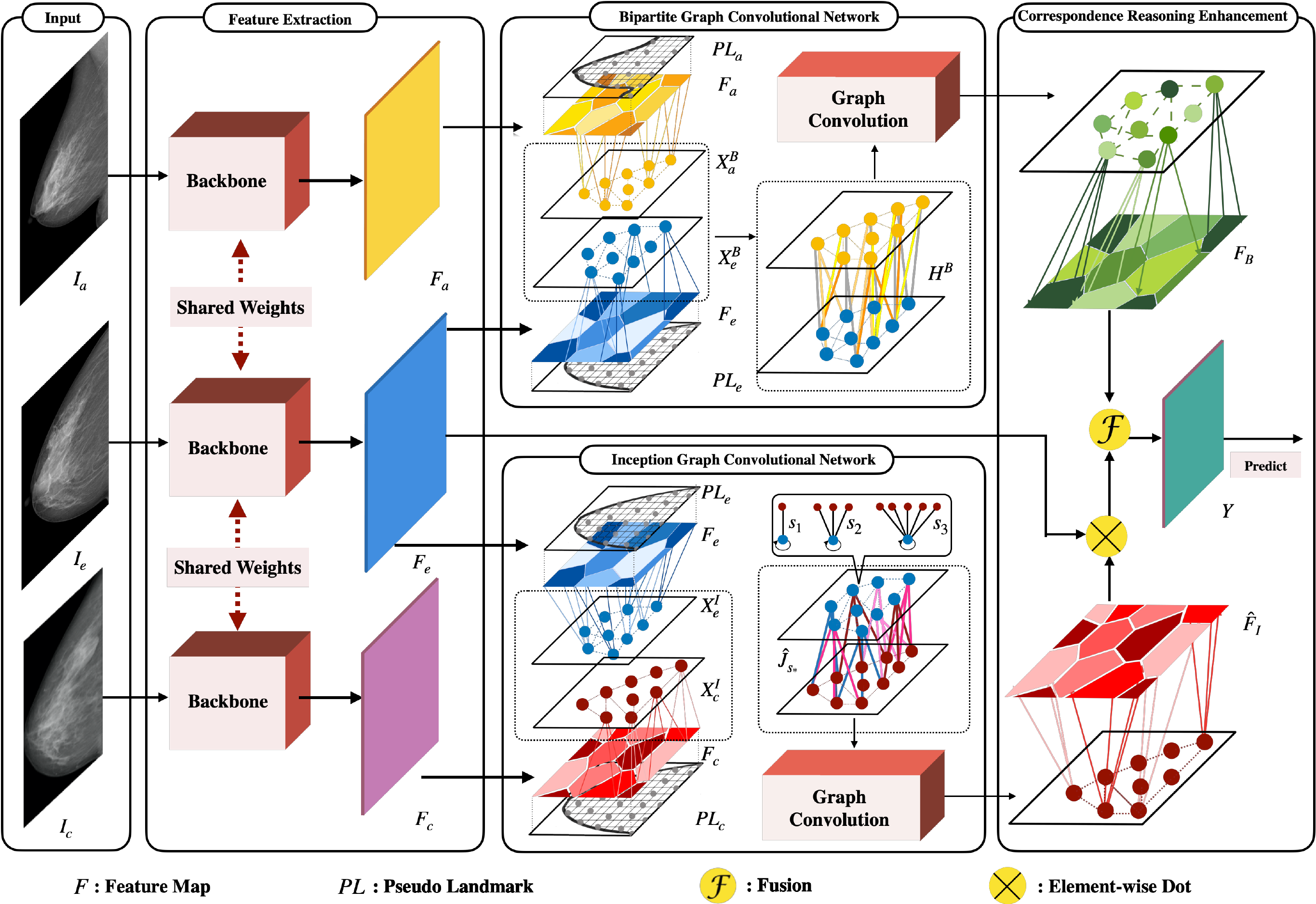}
	\end{center}
	\centering
	\caption{The pipeline of the proposed AGN. AGN takes multi-view backbone features as inputs, and outputs enhanced features of the examined view for further prediction. First, bipartite graph 
		convolutional network performs reasoning across ipsilateral views end outputs auxiliary representations of mass lesion 3D structure. Second, inception graph convolutional network contrasts bilateral views and produces attention maps on the suspicious asymmetric areas. Finally, correspondence reasoning enhancement based on the defined two graphs is conducted to enhance the backbone features of the examined view for further detection.}
	\label{fig:network}
\end{figure*}

\subsection{Multi-view Visual Recognition}
Representing 3D object is one of the most fundamental problems in visual understanding \cite{qi2016volumetric, feng2018gvcnn} and stereo vision \cite{kar2017learning, schops2017multi, Chang_2018_CVPR, chen20173d, rebecq2018emvs}. 
Multi-view based approaches represent the 3D object as a collection of 2D views \cite{su2015multi, johns2016pairwise}. 
Typically, they first conduct image-based classification on each individual view, and then aggregate the multi-view features for 3D representation.
Feng \textit{et al}.~\cite{feng2018gvcnn} propose a group-view CNN for hierarchical correlation modeling from multiple views. 
Yang \textit{et al}.~\cite{yang2019learning} propose a relation network which is capable of modeling region-to-region and view-to-view relationships from different viewpoints. 
For 3D medical images, such as computed tomography (CT) and magnetic resonance imaging (MRI), utilizing 2D image features projected from different views can lead to superior performance \cite{setio2016pulmonary, ciompi2015automatic}. For instance, Setio \textit{et al}.~\cite{setio2016pulmonary} sample multiple views from the 3D CT image, and then cascade different classifiers trained on each individual view to boost the nodule classification performance.

Unlike multi-view based approaches in general computer vision \cite{rebecq2018emvs} and 3D medical image analysis \cite{setio2016pulmonary, ciompi2015automatic}, mammographic views have a differentiated imaging process, which requires us to design customized algorithms for analysis. 
Mammographic views are captured as the total absorption of all substances along the projection ray, which makes it impossible to decomposite the internal structures. 
Ipsilateral views are taken along different directions of the breasts, which provides richer information for representing the 3D structure of breast. 
Meanwhile, bilateral views have similar breast structure, and thus the asymmetric regions between views are more likely to be masses. 
In a nutshell, mammographic views owns more explicit correspondences, which motivates us to develop customized reasoning mechanisms.
We note that the explicit correspondences are also be explored in stereo vision methods~\cite{schops2017multi, Chang_2018_CVPR, chen20173d, rebecq2018emvs}, where they align key points via leveraging the explicit correspondences among calibrated cameras. 
In contrast to stereo vision, we can not obtain precisely matched correspondences between different views due to the existence of standard mammography screening protocols \cite{sampat2005computer}. Thus, how utilize the fuzzy correspondences to enhance the expressive power of backbone features remains an open question.

\section{Preliminary: Mammographic Views}
\label{sec:preliminaries}
In this section, we provide the details of the mammographic screening mechanism. As a special type of 2D radiography, digital mammographic images are captured as the total absorption of all substances along the projection ray. Therefore, using only a single view of mammographic images is insufficient to represent the breast internal structure.  
In standard mammographic screening, X-ray images are taken for both two breasts. For each breast, two mammographic views (i.e., CC view and MLO view) are taken by compressing the breast at a near orthogonal plane. Specifically, CC view is a top-down view while MLO view is a side view taken at a certain angle.
Comparing ipsilateral views  (i.e., both CC and MLO views of the same breast) helps to analyze the 3D structures of masses. Contrasting bilateral views (i.e., a specific view of both breasts) helps to extract suspicious mass lesions since bilateral views of breasts are approximately symmetric.

In this paper, a set of multi-view images are defined as input (illustrated in Figure \ref{fig:motivation}), including an examined view (i.e., the view where the detection is performed), an auxiliary view (i.e., another view of the same breast) and a contralateral view (i.e., the view of opposite side of breast). 
We respectively utilize one of the mammogram images as the examined view, and the rest two views are defined as the auxiliary and contralateral views. 

\section{Methodology}
\label{sec:method}
\subsection{Overview}
The objective of AGN is to endow the mammogram mass detection framework with multi-view correspondence reasoning ability. 
By distilling multi-view information from the input multi-view mammogram images, 
AGN outputs the enhanced feature representations of the examined view for further detection. 
The overall architecture is shown in Figure~\ref{fig:network}, which consists of the following steps.
(1) For the purpose of modeling the region-based reasoning procedure, the graph nodes are embedded into breasts, where each node represents the features of regions with relatively consistent locations in breasts. 
Then, bipartite graph convolutional network is introduced to model the geometric constraints and appearance similarities of nodes between ipsilateral views. 
(2) Inception graph convolutional network is designed to learn structural similarities of bilateral views with an added tolerance of geometric distortions.
(3) Correspondence reasoning enhancement, which is based on the two pre-defined graph convolutional networks, is proposed to enhance the representation power of features. 
Based on the above steps, after information propagated through nodes, each node can not only be aware of the ipsilateral correspondences, but also learn the contrastive representations from bilateral views. 
It is noteworthy that the node representations are mapped to spatial visual domain reversely, which explicitly endows the spatial features with reasoning ability. 
In the end, we fuse the enhanced features and the original backbone features for further proposals. 

Specifically, we are given a set of 2D feature maps $F_{e}, F_{a}, {F_{c} \in \mathbb{R} ^ {HW \times C}}$ extracted from the examined view (simplified as $e$), the auxiliary view (simplified as $a$) and the contralateral view (simplified as $c$),  and $H, W$ and $C$ represent the height, width and channel of the feature maps. $l_e, l_a, l_c \in \{CC, MLO\}$ are defined as view types.  We guarantee that $l_e \neq l_a$ and $l_e = l_c$. 
As formulated in Equation \ref{eq_overall},  AGN learns a function $f$, parameterized by the bipartite graph $\mathcal{G}_B$ and the inception graph $\mathcal{G}_I$.
\begin{equation}
	\label{eq_overall}
	\begin{aligned}
		Y = f(F_{e}, F_{a}, F_{c}; \mathcal{G}_B, \mathcal{G}_I)
	\end{aligned}
\end{equation}

\subsection{Graph Nodes}
Graph nodes are introduced to denote the region-level correspondences in breasts with relatively consistent locations across different breast instances. Technically, we define graph nodes by considering two fundamental questions, i.e., ``where to locate'' and `` what to represent''.

We introduce the concept of pseudo landmarks, which preserve relative consistent locations in breasts, to address the first question. For the second question, we note that the graph node mapping can produce node representations from spatial visual features. 
In the following parts, we provide the technical details.

\subsubsection{Pseudo Landmarks}
Landmarks represents points in a shape object, where correspondences between and within the populations of the object are preserved~\cite{Drydmard16}. 
Unfortunately, there are no specialized landmarks for breasts, which inspires us to define pseudo landmarks based on prior knowledge. 

The pseudo landmarks are expected to possess the following properties: \uppercase\expandafter{\romannumeral1.} Each pseudo landmark stands for a region with relatively consistent locations in breasts; \uppercase\expandafter{\romannumeral2.} Different pseudo landmarks should stand for distinct regions in breasts; \uppercase\expandafter{\romannumeral3.} Combining all pseudo landmarks are expected to cover the whole breast.

An intuitive approach is to regard uniform grids of the image as landmarks. However, property \uppercase\expandafter{\romannumeral1.} is not satisfied since the uniform grids are sensitive to the variations of image scale, geometric structures, etc. 
As demonstrated in Figure \ref{fig:motivation}, the design principle of pseudo landmarks is based on a key observation: there exist clear geometric correspondences between CC and MLO views of standard mammography screening.
Ideally, a point in CC view approximately corresponds to a line in MLO view, which is parallel to the projected pectoral muscle plane.

\begin{figure}[t]
	\centering
	\subfigure[CC View]{\label{fig:1}\includegraphics[height=0.57\linewidth]{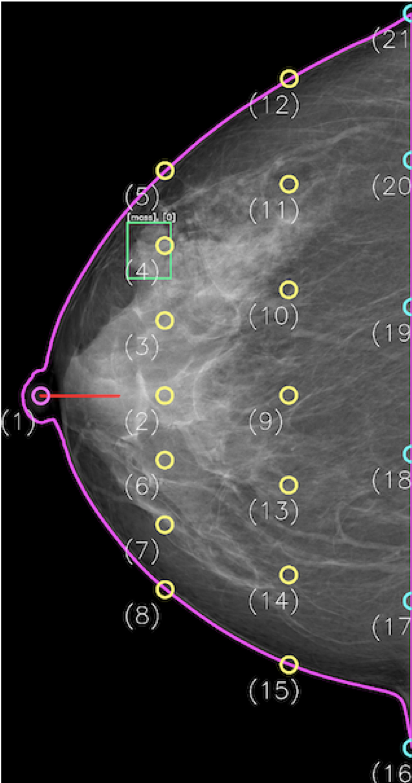}}
	\subfigure[MLO View]{\label{fig:2}\includegraphics[height=0.57\linewidth]{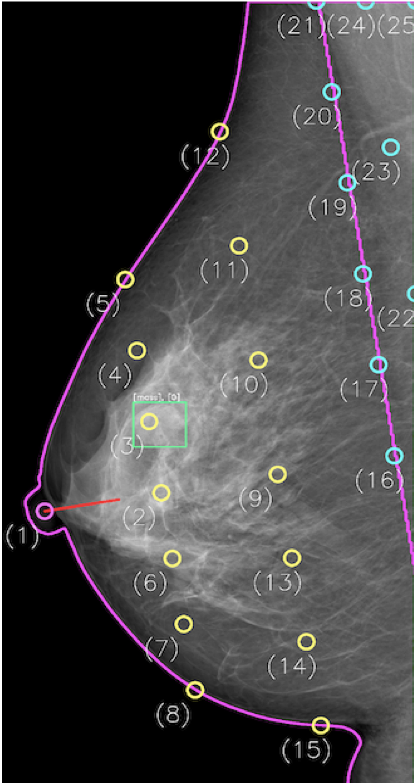}}
	\subfigure[Mapping Cell]{\label{fig:3}\includegraphics[height=0.57\linewidth]{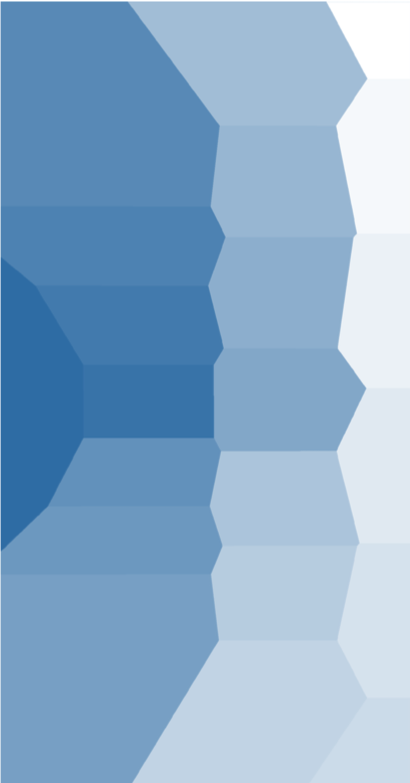}}
	\caption{Illustration of pseudo landmarks and bipartite graph node mapping. (a)-(b) draw pseudo landmarks and the matched bounding  boxes on CC and MLO views respectively. 
		(c) illustrates how bipartite node mapping works when $k=1$. 
		Each mapping cell denotes the representative region of the node in the CC view.
	}
	\label{fig:pseudo_landmark}
\end{figure}

As illustrated in Figure \ref{fig:pseudo_landmark}, in order to embed the pseudo landmarks, we first insert a set of equidistant parallel lines between the nipple and pectoral muscle line (projected by the pectoral muscle plane). Based on the intersection between parallel lines and breast contour, we uniformly insert points between two intersection points. After that, all the points are re-ordered based on the intersections and further defined as pseudo landmarks. Similarly, we define pseudo landmarks in pectoral muscle areas for MLO view. By doing so, we can obtain a set of pseudo landmarks for each view.

\subsubsection{Graph Node Mapping}
\label{chapter: node_mapping}
Graph node mapping targets on projecting spatial visual features ${F} \in \mathbb{R} ^ {HW \times C} $ to the node domain. Note that the features of each node are region-level features that represent a certain region in breasts.

The node mapping denotes the relation between a graph node and all pixels in the corresponding region. 
Formally, we design kNN ($k$ Nearest Neighbor) forward mapping $\phi_k$ with its auxiliary matrix $A$ for node feature representations. 
Each node is associated to an irregular region, satisfying the property that for any pixel in this region, the node is one of its $k$ nearest nodes.
$\phi_k$ performs region-level feature pooling within the regions corresponding to the graph nodes. The formulas are defined as follows:
\begin{equation}
	\begin{aligned}
		\phi_k({F}, \mathcal{V}) = ({Q}^f)^{T} {F},
	\end{aligned}
\end{equation}
\begin{equation}
	\begin{aligned}
		{Q}^{f} = {A} ({\Lambda}^{f})^{-1},
	\end{aligned}
\end{equation}
\begin{equation}
	\label{assign_matrix}
	\begin{aligned}
		A_{ij}  = 
		\begin{cases}
			1, & \text{if $j$ th node is kNN of $i$ th pixel.} \\
			0, & \text{otherwise.}
		\end{cases}
	\end{aligned}
\end{equation}
where $\mathcal{V}$ denotes the node set corresponding to spatial feature ${F} \in \mathbb{R} ^ {HW \times C}$, ${A} \in \mathbb{R} ^ {HW \times |\mathcal{V} |}$  is the auxiliary matrix that assigns spatial features to top-$k$ nearest graph nodes, ${\Lambda}^{f} \in \mathbb{R}^{\mathcal{|\mathcal{V}| \times |\mathcal{V}|}} $ ($\Lambda^{f}_{jj} = \sum\limits_{i=1}^{HW} A_{ij}$) is a diagonal matrix, and ${Q} ^{f} \in \mathbb{R} ^ {HW \times |\mathcal{V}|}$ (a normalized form of $A$) denotes the forward mapping matrix. 

Compared to fixed-grid assign methods~\cite{gao2019graph},
the node representations in our AGN are more robust to the variations of image scales, geometric structures, etc. 
The justification is that $\phi_k$ adaptively chooses representative region according to the relations among node locations. 
In addition, the proposed mapping mechanism has an explicit physical meaning, which has the merit of better visual interpretability. 
In particular, as show in figure \ref{fig:pseudo_landmark}~(c), the mapping degenerates to Voronoi grids \cite{aurenhammer2000voronoi} when $k=1$. 

\subsection{Bipartite Graph Convolutional Network (BGN)}
BGN learns to model ipsilateral relations among correspondences. BGN is characterized as $\mathcal{G}_B = (\mathcal{V}_{CC}, \mathcal{V}_{MLO}, \mathcal{E}_{B} )$, where $\mathcal{V}_{CC}$ and $\mathcal{V}_{MLO}$ represent the bipartite graph node sets constructed from CC view and MLO view respectively. $\mathcal{E}_{B}$ denotes the bipartite graph edge set. Each edge in $\mathcal{E}_{B}$ connects a node in $\mathcal{V}_{CC}$ to the corresponding one in $\mathcal{V}_{MLO}$, leading to a bipartite graph structure. 

To obtain the feature representations of bipartite graph node, we resort to the kNN forward mapping described in Section \ref{chapter: node_mapping}. Formally, the bipartite graph node representations are defined as follows:
\begin{equation}
	\begin{aligned}
		X_e^B = \phi_k(F_{e}, \mathcal{V}_{l_{e}}),
	\end{aligned}
\end{equation}
\begin{equation}
	\begin{aligned}
		X_a^B = \phi_k(F_{a}, \mathcal{V}_{l_{a}}).
	\end{aligned}
\end{equation}
Here, for the sake of simplicity, we denote $X^{CC} \in \{X_e^B, X_a^B\}$ as node features for CC view and $X^{MLO} \in \{X_e^B, X_a^B\}$ for MLO view.

To obtain bipartite graph edge representations, we start by rethinking a fundamental question: what is the underlying relations between nodes? 
Given a mass which is located at one certain node in the examined view, it is clear that different nodes in the auxiliary view will have differentiated probabilities for representing the given mass.
Motivated by this, we regularize the relation from two aspects, i.e., geometric constraints and appearance similarities. The two aspects characterize the inherent constraints resulted from mammogram screening mechanism and visual similarities between nodes, respectively. 

Formally, bipartite graph edge is denoted as an adjacency matrix ${H} \in \mathbb{R}^{|\mathcal{V_{CC}}| \times |\mathcal{V}_{MLO}|}$, which consists of a geometric graph  ${H}^g \in \mathbb{R}^{|\mathcal{V}_{CC}| \times |\mathcal{V}_{MLO}|}$ and a semantic graph ${H}^s \in \mathbb{R}^{|\mathcal{V}_{CC}| \times |\mathcal{V}_{MLO}|}$. The geometric graph is a global prior graph, which represents the geometric constraints across views. The semantic graph is an instance dependent graph, which characterizes the semantic similarities between nodes. 
The two graphs jointly regularize the ipsilateral information propagation. 
Eq.~\eqref{bipartite_graph} demonstrates the relations of these two matrices, 
\begin{equation} 
	\label{bipartite_graph}
	\begin{aligned}
		{H} ={H}^g \circ {H}^s
	\end{aligned}
\end{equation}
where $\circ$ denotes the element-wise dot. 

In the following subsections, we show the details of how to regularize the relations among correspondences.

\subsubsection{Geometric Relation Learning}
In this part, we explicitly model the geometric constraints.   
Even though the CC and MLO views own standard camera pose, it is difficult to define the precise geometric correspondence due to the tissue deformation and the lack of visual cues. 
To solve this issue, we resort to use masses as visual cues for modeling the geometric correspondence.
Each edge in the geometric graph stands for the correlation of the linked nodes (i.e., the same mass instance from different views).
In order to estimate the correlation, for each mass, if a node is the closest to the center of bounding box, this node will be selected to represent the mass. By doing so, we are capable of linking the nodes that stands for the same mass instance from different views (e.g., 4$^{th}$ node in CC view and 3$^{th}$ node in MLO view in Figure \ref{fig:pseudo_landmark}).

The construction of geometric graph $H^g$ consists of the following two steps.
(1) We construct a frequent statistics matrix $\epsilon$ for masses. If a node is the closest one to the center of the bounding box of a mass, this node is chosen to represent the mass. By doing so, we are capable of linking the nodes that stand for the same mass instance from different views. We traverse all labeled masses in the training set and obtain the frequent statistics matrix $\epsilon \in  \mathbb{R} ^{|\mathcal{V}_{CC}| \times |\mathcal{V}_{MLO} |}$.	
(2) To obtain $H^g$, we perform an augmented form of column-row normalization~\cite{xu2019reasoning}:
\begin{equation}
	H_{ij}^g=\frac{\epsilon_{ij}}{\sqrt{D_{i\cdot}D_{\cdot{j}}}}
\end{equation}
where $D_{i\cdot}=\sum\limits_{k=1}\epsilon_{ik}$ and $D_{\cdot{j}}=\sum\limits_{k=1}\epsilon_{kj}$.

\subsubsection{Semantic Relation Learning}
Whilst the geometric graph characterize the holistic geometric correlations, it still inevitably introduces noises during the reasoning procedure, and thus is hard to find exact correspondence pairs across views. In addition, we note that the appearance similarities between different views is a significant characteristic of mass lesions. Inspired by this, we introduce the semantic graph to learn the semantic relation between nodes, which is helpful for mitigating the negative influence of the noisy relations.

An intuitive approach to define the semantic similarities between nodes is to measure them by cosine similarity or inner product~\cite{tomasi1998bilateral,buades2005non}. However, the  relations between nodes include the backgrounds, and their features may also be enhanced.
To tackle this issue, we relax the weights, and allow the module to learn its own similarity as follows:
\begin{equation}
	\begin{aligned}
		{H}_{ij}^{s} = \sigma( 
		[ (X_i^{CC})^T, (X_j^{MLO})^T ] w_s),
	\end{aligned}
\end{equation}
where $X_i^{CC}, X_j^{MLO} \in \mathbb{R}^{C} $ respectively denote the $i^{th}$ and $j^{th}$ node features of $CC$ and $MLO$ views, $w_s \in \mathbb{R}^{2C}$ stands for the fusion parameter, and $\sigma$ denotes the sigmoid activation function.  

\subsection{Inception Graph Convolutional Network (IGN)}
\label{sec: ign}
Based on the assumption that bilateral mammogram views share a similar breast structure and asymmetric regions are more likely to be suspicious regions, we propose the IGN to learn to contrast bilateral mammogram views.
IGN links nodes with compatible locations from bilateral views and predicts attention values for regions in the examined view.

IGN is characterized as $\mathcal{G}_I = (\mathcal{V}_{e} \cup \mathcal{V}_{c}, \mathcal{E}_{I})$, where $\mathcal{V}_{e}, \mathcal{V}_{c}$ indicate node sets constructed from the examined view and the contralateral view respectively. 
Since bilateral views have the same view type (i.e., $l_e = l_c$), we guarantee that $|\mathcal{V}_{e}| = |\mathcal{V}_{c}|$. 
For simplicity, we assume that $n = |\mathcal{V}_{e}| = |\mathcal{V}_{c}|$.

To obtain node feature representations for IGN, we adopt kNN forward mapping similarly. Formally, representations corresponding to the examined view and the contralateral view are defined as:
\begin{equation}
	\begin{aligned}
		X_e^I = \phi_k(F_{a}, \mathcal{V}_{e})
	\end{aligned}
\end{equation}
\begin{equation}
	\begin{aligned}
		X_c^I = \phi_k(F_{c}, \mathcal{V}_{c})
	\end{aligned}
\end{equation}
Then, the node representation of IGN can be defined as:
\begin{equation}
	\begin{aligned}
		X^I = [(X_e^I)^T, (X_c^I)^T]^T
	\end{aligned}
\end{equation}

To obtain edge representation of IGN, we characterize the edge set $\mathcal{E}_{I}$ as an 
adjacency matrix $\hat{J} = \left(\begin{matrix}
	M & J  \\
	J^T & M^T\\
\end{matrix}\right)$, which contains two components, i.e., $M \in \mathbb{R} ^ {n \times n}$ and $J \in \mathbb{R} ^ {n \times n}$. 
Specifically, $M$ characterizes the relations of nodes within the same view, while $J$ characterizes the relations of nodes across different views. 

We set $M$ to $\mathbf{0}$, indicating that there are no intra-connections within the view. However, determining the attention values of a certain view not only requires its contralateral information  (i.e. $J$) but also the view information itself. Thus, we add self-loop for each node of the graph. Specifically, we set $M=I_n$. 

As for the definition of $J$,
it is intuitive to set $J = I_n$ which assumes that only nodes with the same location in bilateral views are linked. However, the bilateral views may not be aligned perfectly due to the inherent geometric distortions. To tolerate the distortions, we reformulate $J$ as $J_s$ which links each node to its top-\textit{s} nearest neighbors (NN) in the contralateral view. $J_s$ provides larger visual context which helps to increase the tolerance for the distortions.
Note that each distinct value of $s$ can induce a distinct set of cross-view edges and their corresponding cross-view adjacency matrix is denoted as $J_s$. 

The set of cross-view edges corresponding to $J_s$ can be regarded as a single branch of graph linkages. Instead of using a single branch, we adopt multiple branches of graph linkages, each corresponding to a distinct cross-view adjacency matrices. This kind of GCN can provide stronger representation abilities and form an Inception-like structure \cite{szegedy2015going}. 
Specifically, supposing that IGN has two different branches $s_1, s_2 \in \mathbb{N}^*$, the induced corresponding augmented adjacency matrices are denoted as $\hat{J}_{s1}, \hat{J}_{s2}$. When performing graph convolutions, both $\hat{J}_{s_{1}}$ and $\hat{J}_{s_{2}}$ affect information propagation. Details of convolution operation will be described in Section \ref{sec: inception_conv}.

\subsection{Correspondence Reasoning Enhancement}
\label{sec: reasoning}
Correspondence reasoning  enhancement, which is based on the defined bipartite graph $\mathcal{G}_B$ and the inception graph $\mathcal{G}_I$, is developed to fully explore multi-view reasoning procedure for enhancing the customized features. It consists of the following steps.
(1) Augment bipartite graph convolution, making it adapt to the modern graph convolutional manner; 
(2) Design inception graph convolution with multi-branch connections among nodes;
(3) Map node representations to spatial domain reversely after several layers of graph convolutions;
and (4) Fuse the original backbone features with the graph representations learned from BGN and IGN to enhance the expressive power of final representations.  
We will describe the details in the following subsections.

\subsubsection {Bipartite Graph Convolution}
To adapt the modern GCN \cite{gao2019graph, kipf2016semi}, we provide the augmented form of the bipartite graph:
\begin{equation}
	\begin{aligned}
		X^B = [(X^{CC})^T, (X^{MLO})^T]^T
	\end{aligned},
\end{equation}
\begin{equation}
	\begin{aligned}
		H^B =
		\left(
		\begin{matrix}
			\mathbf{0} & H  \\
			H^T & \mathbf{0}\\
		\end{matrix}
		\right),
	\end{aligned}
\end{equation}
where $X^B \in \mathbb{R}^{|\mathcal{V}_{CC} \cup \mathcal{V}_{MLO} | \times C}$ denotes the augmented form of the bipartite graph nodes, and $H^B \in \mathbb{R}^{|\mathcal{V}_{CC} \cup \mathcal{V}_{MLO} | \times  |\mathcal{V}_{CC} \cup \mathcal{V}_{MLO} |}$ is the augmented form of the adjacency matrix.

We follow the common practice~\cite{gao2019graph} to define graph convolution. An iteration of graph convolution layer is defined in Equation \ref{gcn}, where  $W^{B} \in \mathbb{R} ^ {C \times C}$ and $\sigma$ indicate the convolution parameters and sigmoid activation function. To this end, we are able to stack multiple layers to form the the graph convolutional network. 
\begin{equation}
	\label{gcn}
	\begin{aligned}
		Z^B = \sigma( H^B X^B W^{B})
	\end{aligned}
\end{equation}

\subsubsection {Inception Graph Convolution}
\label{sec: inception_conv}

To achieve inception graph convolution with multi-branch linkage among nodes, we generalize standard graph convolution operations. We give the formulation of an iteration of the inception graph convolutional operation in Equation \ref{eq:multi_scale_conv}. Feature transformations of multiple branches are conducted independently, and then the transformed multi-branch features are aggregated. For simplicity, the equation contains only two branches. It is intuitive to reformulate it to adapt to multi-branch settings. 
\begin{equation}
	\label{eq:multi_scale_conv}
	\begin{aligned}
		Z^I = \sigma\left(
		\left(\begin{matrix}
			\hat{J}_{s_1} \hat{J}_{s_2}
		\end{matrix}\right)
		\left(\begin{matrix}
			X^I & \mathbf{0} \\
			\mathbf{0} & X^I \\
		\end{matrix}\right)
		\left(\begin{matrix}
			W^I_1 \\
			W^I_2 \\
		\end{matrix}\right)
		\right),
	\end{aligned}
\end{equation}
where $W^I_1, W^I_2 \in \mathbb{R} ^ {C \times C}$ indicate parameters of the layer.

\subsubsection{kNN Reverse Mapping}
In order to enhance the spatial features, we introduce a kNN reverse mapping function $\psi_{k}$ to map the graph node features to the spatial domain.
Following the design principle of the kNN forward mapping (cf. Section \ref{chapter: node_mapping}), we keep the same number ($k$) of nearest neighbors.
$\psi_{k}$ is defined as :
\begin{equation}
	\begin{aligned}
		\psi_k(Z, \mathcal{V}_e) = Q^r [Z]_e,
	\end{aligned}
\end{equation}
\begin{equation}
	\begin{aligned}
		Q^{r} = (\Lambda^{r})^{-1} A,
	\end{aligned}
\end{equation}
where $Z$ denotes the node presentations after graph convolutions,  $\mathcal{V}_e$ stands for the node set from the examined view,  $A \in \mathbb{R} ^ {HW \times |\mathcal{V}_e |}$, which is correspond to $\mathcal{V}_e$, is defined by following Equation \ref{assign_matrix}, $[\cdot]_e$ denotes an indexing operator which chooses nodes in the examined view from all nodes, $\Lambda^{r} \in \mathbb{R}^{HW \times HW} $ represents a diagonal matrix, $\Lambda^{r}_{ii} = \sum\limits_{j=1}^{|\mathcal{V}_e|} A_{ij}$, and $Q ^{r} \in \mathbb{R} ^ {HW \times |\mathcal{V}_e|}$ denotes the reverse mapping matrix which is the normalized form of $A$.

\subsubsection{Feature fusion}
Feature fusion procedure includes the following steps. We first map the features from the auxiliary view to spatial domain using $\psi_k$. After that, $F_B$ and $F_e$ are aligned to the same coordinate space, which helps to fuse them more effectively:
\begin{equation}
	\begin{aligned}
		F_B  = \psi_k(Z^B, \mathcal{V}_e)
	\end{aligned}
\end{equation}
Then, we predict attention values for regions in the examined view induced by inception graph convolutions.
\begin{equation}
	\begin{aligned}
		F_I = \psi_k(Z^I, \mathcal{V}_e),
	\end{aligned}
\end{equation}
\begin{equation}
	\begin{aligned}
		\hat{F}_I = \sigma(F_I w_I),
	\end{aligned}
\end{equation}
where $w_I \in \mathbb{R} ^ {C}$ represents the parameter, and $\hat{F}_I$ refers to the spatial attention map. Finally, we enhance the features based on the processed feature triple, which has been aligned in the same spatial coordinate space:
\begin{equation}
	\begin{aligned}
		Y =[\hat{F}_I \cdot F_e, F_B] W_f^T,
	\end{aligned}
\end{equation}
where $\cdot$ indicates spatial-wise dot which broadcasts along channel axis, $W_f \in \mathbb{R} ^ {C \times 2C}$ represents the fusion parameter.

\begin{table}[t]
	\caption{Performance  on DDSM dataset(\%).}
	\label{ddsm_report}
	\centering
	\footnotesize 
	\renewcommand{\arraystretch}{1.2}
	\centering
	\label{experiment_ablation}
	\begin{tabular}{p{3cm}<{\centering}  p{4cm}<{\centering} p{0.6cm}<{\centering} p{0.6cm}<{\centering} p{0.6cm}<{\centering} p{0.6cm}<{\centering} p{0.6cm}<{\centering} p{0.6cm}<{\centering}}
		\hline
		\textbf{Method} & R@t \\
		\hline
		Campanini~\textit{et at}.~\cite{campanini2004novel} & 80@1.1\\
		Eltonsy~\textit{et at}.~\cite{eltonsy2007concentric} & 92@5.4, 88@2.4, 81@0.6\\
		Sampat~\textit{et at}.~\cite{sampat2008model} & 88@2.7, 85@1.5, 80@1.0\\
		Faster RCNN ~\cite{ma2019cross} & 85@2.1, 75@1.8, 73@1.2\\
		CVR-RCNN ~\cite{ma2019cross} & 92@4.4, 88@1.9, 85@1.2\\
		\hline
		\textbf{AG-RCNN} & \textbf{96@4.4}, \textbf{92@1.9}, \textbf{90@1.2}\\
		\hline
	\end{tabular}
\end{table}

\begin{table}[t]
	\caption{Performance on DDSM dataset(\%).}
	\label{ddsm_our}
	\centering
	\footnotesize 
	\renewcommand{\arraystretch}{1.2}
	\centering
	\label{experiment_ablation}
	\begin{tabular}{p{3.2cm}<{\centering}  p{0.57cm}<{\centering} p{0.57cm}<{\centering} p{0.57cm}<{\centering} p{0.57cm}<{\centering} p{0.57cm}<{\centering} p{0.57cm}<{\centering} p{0.57cm}<{\centering}}
		\hline
		\textbf{Method} & R@0.5 & R@1.0 & R@2.0 & R@3.0 & R@4.0 \\
		\hline
		Faster RCNN, FPN & 75.3 & 81.5 & 87.3 & 89.8 & 91.4\\
		Faster RCNN, FPN, DCN  & 75.7 & 82.5 & 88.4 & 90.1 & 91.4\\
		Mask RCNN, FPN & 76.0 & 82.5 & 88.7 & 90.8 & 91.4\\
		Mask RCNN, FPN, DCN & 76.7 & 83.9 & 89.4 & 91.4 & 91.8\\
		BG-RCNN \cite{Liu_2020_CVPR} & 79.5 & 86.6 & 91.8 & 92.5  & 94.5\\
		\hline
		\textbf{AG-RCNN} & \textbf{82.0} & \textbf{89.0} & \textbf{92.1} & \textbf{93.8}  & \textbf{95.5}\\
		\hline
	\end{tabular}
\end{table}

\section{Experiments}
\label{sec: experiment}

\subsection{Implementation Details}
\label{implementation_details}
In experiments, the mammogram images are segmented by OTSU~\cite{4310076}, and we use the foreground regions as the input. 
In order to keep the same spatial resolution among different view of images, each view of input image is resized to same size of the examined image.
We leverage the Hough transform to detect the pectoral muscle line and nipple in three steps for pseudo landmark embedding. 
First, points potentially lying on the pectoral muscle line are extracted with the Canny edge detector. Then, these extracted points are mapped into the parameter space. Finally, noisy points in the parameter space are removed according to the prior location of the pectoral muscle line, and the optimal point in the parameter space is identified for the pectoral muscle line. The point on the breast contour with the largest distance to the detected pectoral muscle line is further located as the nipple.
We apply several specific data augmentation methods, such as random flipping, random cropping, and multi-scaling, to prevent over-fitting during the training stage.

The proposed AGN is integrated into Mask RCNN~\cite{he2017mask} with ResNet-50~\cite{he2016deep} architecture, and we term the full mammogram mass detection framework as AG-RCNN hereafter. 
The parameters of ResNet-50 are fine-tuned from the model pre-trained on ImageNet.
The loss function follows the same definition as Mask RCNN \cite{he2017mask}, containing three terms: classification loss, regression loss, and segmentation loss.
The FPN anchors span 5 scales and 3 aspect ratios \cite{lin2017feature}.
To adapt to modern FPN \cite{lin2017feature} network structure, AGN enhances each level of feature pyramid with shared parameters, since node representations are invariant to feature map scales.

Our experiments are implemented by the PyTorch deep learning framework \cite{dhungel2015automated}. 
We utlize stochastic gradient descent (SGD) for the training with a learning rate $0.02$, weight decay $10^{-4}$, momentum $0.9$ and nesterov set True. The training process takes 30 epochs in all. 
Regarding the BGN, we keep the same number of nearest neighbors $k$ for both $\phi_k$ and $\psi_k$ for bipartite node mapping and reverse mapping. As for IGN, the number of nearest neighbors $k$ for both $\phi_k$ and $\psi_k$ is set to 1 for keeping higher spatial resolutions.

\subsection{Datasets}
\label{Datasets}
We perform experiments on both a public dataset DDSM \cite{heath2000digital} and an in-house dataset. Note that we do not choose other public datasets (such as INBreast \cite{moreira2012inbreast}
, MIAS \cite{suckling1994mammographic}). The justification is that the sample size of these datasets is insufficient to train a detection model. 
\\

\noindent \textbf{DDSM dataset.} 
DDSM dataset includes 2620 mammography cases, and most cases contain two views of images for both breasts. Following previous practices~\cite{ma2019cross, eltonsy2007concentric, campanini2004novel, sampat2008model}, the DDSM dataset is divided to 1897 cases for training, 211 cases for validation and 512 cases for testing.
\\

\noindent \textbf{In-house dataset. } 
The in-house dataset includes 10,000 cases, which are collected from four different vendors: IMS s.r.l., Siemens, Hologic, and GE Healthcare. Each case contains a CC view and an MLO view for each breast, and thus there are 40,000 images in all. The annotations, namely the mask of each mass lesion, are labeled by 3 radiologists with strong expertise. 
If there are disagreements among them, we will adopt the majority opinion of radiologists. This dataset is randomly split into training, validation and testing sets in a ratio of 8:1:1.

\begin{table}[t]
	\caption{Performance on in-house dataset(\%).}
	\label{in_house}
	\centering
	\footnotesize 
	\renewcommand{\arraystretch}{1.2}
	\centering
	\label{experiment_ablation}
	\begin{tabular}{p{3.2cm}<{\centering}  p{0.57cm}<{\centering} p{0.57cm}<{\centering} p{0.57cm}<{\centering} p{0.57cm}<{\centering} p{0.57cm}<{\centering} p{0.57cm}<{\centering} p{0.57cm}<{\centering}}
		\hline
		\textbf{Method} & R@0.5 & R@1.0 & R@2.0 & R@3.0 & R@4.0 \\
		\hline
		Faster RCNN, FPN & 82.3 & 85.4 & 90.5 & 92.5	 & 93.7\\
		Faster RCNN, FPN, DCN  & 83.1 & 88.0 & 91.0 & 92.5 & 93.9\\
		Mask RCNN, FPN & 83.1 & 88.0 & 91.4 & 93.4 & 94.2\\
		Mask RCNN, FPN, DCN & 84.0 & 88.3 & 91.7 & 93.2 & 94.5\\
		BG-RCNN \cite{Liu_2020_CVPR} & 85.7 & 89.4 & 92.1 & 93.7 & 95.0\\
		\hline
		\textbf{AG-RCNN} & \textbf{87.6} & \textbf{90.6} & \textbf{93.4} & \textbf{94.7}  & \textbf{95.2}\\ 
		\hline
	\end{tabular}
\end{table}

\subsection{Baselines}
\label{Baselines}

\noindent \textbf{Faster RCNN, FPN. } \ 
Faster RCNN \cite{ren2015faster} with Feature Pyramid Network (FPN) \cite{lin2017feature} is a strong baseline in object detection task. FPN enhances the modeling ability of detecting multi-scale objects. It assigns objects to different level of feature maps according to object scales. We use ResNet-50\cite{he2016deep} as the backbone network.
\\

\noindent \textbf{Faster RCNN, FPN, DCN. }  \
Deformable Convolution Network (DCN) \cite{dai2017deformable} is proposed to boost the transformation modeling capability of convolutional networks. DCN is introduced into the baselines to further enhance the detection performance. 
\\

\noindent \textbf{Mask RCNN, FPN, DCN. }  \ 
Mask RCNN \cite{he2017mask} is a state-of-the-art approach for both object detection and instance segmentation tasks. To leverage mask annotations for precise localization, we exploit Mask RCNN framework for boosting the performance.
\\

\begin{table*}[]
	\caption{Effectiveness of pseudo landmarks on  DDSM and In-house datasets (\%).}
	\label{ablation_landmark}
	\centering
	\footnotesize
	\renewcommand{\arraystretch}{1.2}
	\begin{tabular}{c|ccccc|ccccc}
		\hline
		\multicolumn{1}{c|}{\multirow{2}{*}{\textbf{Method}}} & \multicolumn{5}{c|}{\textbf{DDSM}} & \multicolumn{5}{c}{\textbf{In-house}} \\ \cline{2-11}
		\multicolumn{1}{c|}{} & \multicolumn{1}{c}{\textbf{R@0.5}} & \multicolumn{1}{c}{\textbf{R@1.0}} & \multicolumn{1}{c}{\textbf{R@2.0}} & \multicolumn{1}{c}{\textbf{R@3.0}} &  \multicolumn{1}{c|}{\textbf{R@4.0}} & \multicolumn{1}{c}{\textbf{R@0.5}} & \multicolumn{1}{c}{\textbf{R@1.0}} & \multicolumn{1}{c}{\textbf{R@2.0}} & \multicolumn{1}{c}{\textbf{R@3.0}} & \textbf{R@4.0} \\ \hline
		Uniform Grids  &  77.7 & 85.6& 91.5 & 93.2 & 94.5 &  85.6 & 89.4& 81.9 & 93.0 & 94.1\\
		\textbf{Proposed} &  \textbf{82.0} & \textbf{89.0} & \textbf{92.1} & \textbf{93.8}  & \textbf{95.5} & \textbf{87.6} & \textbf{90.6} & \textbf{93.4} & \textbf{94.7}  & \textbf{95.2}\\
		\hline
	\end{tabular}
\end{table*}

\begin{table*}[]
	\caption{Effectiveness of node number on  DDSM and In-house datasets (\%).}
	\label{ablation_nb_node}
	\centering
	\footnotesize
	\renewcommand{\arraystretch}{1.2}
	\begin{tabular}{c|ccccc|ccccc}
		\hline
		\multicolumn{1}{c|}{\multirow{2}{*}{\textbf{Method}}} & \multicolumn{5}{c|}{\textbf{DDSM}} & \multicolumn{5}{c}{\textbf{In-house}} \\ \cline{2-11}
		\multicolumn{1}{c|}{} & \multicolumn{1}{c}{\textbf{R@0.5}} & \multicolumn{1}{c}{\textbf{R@1.0}} & \multicolumn{1}{c}{\textbf{R@2.0}} & \multicolumn{1}{c}{\textbf{R@3.0}} &  \multicolumn{1}{c|}{\textbf{R@4.0}} & \multicolumn{1}{c}{\textbf{R@0.5}} & \multicolumn{1}{c}{\textbf{R@1.0}} & \multicolumn{1}{c}{\textbf{R@2.0}} & \multicolumn{1}{c}{\textbf{R@3.0}} & \textbf{R@4.0} \\ \hline
		$PL(1, 1)$ &  77.0 & 84.2 & 91.4 & 92.5 & 93.2 & 84.5 & 89.6 & 92.1 & 94.2 & 94.7\\
		$PL(9, 13)$ &  80.5 & 87.1 & 91.6 & 93.5 & 94.5 &  86.4 & 90.1 & 92.6 & 94.5 & 95.0 \\ 
		$PL(21, 25)$ & 81.1 & 87.0 & 91.8 & 93.2 & 94.5 & 87.2 & 90.0 & 92.9 & 94.3 & 95.0\\
		$PL(42, 46)$ & 81.4 & 87.3 & 92.0 & 93.8 & 95.2 & 87.1 & 90.4 & 93.2 & \textbf{94.9} & 95.2\\
		$PL(66, 71)$ & \textbf{82.0} & \textbf{89.0} & \textbf{92.1} & \textbf{93.8} & \textbf{95.5} & \textbf{87.6} & \textbf{90.6} & \textbf{93.4} & 94.7 & \textbf{95.2}\\
		\hline
	\end{tabular}
\end{table*}

\begin{table*}[]
	\caption{Effectiveness of graph node mapping on  DDSM and In-house datasets (\%).}
	\label{ablation_bipartite_mapping}
	\centering
	\footnotesize
	\renewcommand{\arraystretch}{1.2}
	\begin{tabular}{c|ccccc|ccccc}
		\hline
		\multicolumn{1}{c|}{\multirow{2}{*}{\textbf{Method}}} & \multicolumn{5}{c|}{\textbf{DDSM}} & \multicolumn{5}{c}{\textbf{In-house}} \\ \cline{2-11}
		\multicolumn{1}{c|}{} & \multicolumn{1}{c}{\textbf{R@0.5}} & \multicolumn{1}{c}{\textbf{R@1.0}} & \multicolumn{1}{c}{\textbf{R@2.0}} & \multicolumn{1}{c}{\textbf{R@3.0}} &  \multicolumn{1}{c|}{\textbf{R@4.0}} & \multicolumn{1}{c}{\textbf{R@0.5}} & \multicolumn{1}{c}{\textbf{R@1.0}} & \multicolumn{1}{c}{\textbf{R@2.0}} & \multicolumn{1}{c}{\textbf{R@3.0}} & \textbf{R@4.0} \\ \hline
		$PL(21, 25)$, Crop &  78.8 & 86.3 & 91.1 & 92.5 & 94.2 & 85.4 & 88.4 & 91.4 & 93.2 & 94.2 \\
		$PL(21, 25)$, $k=1$ & 80.8 & 86.8 & 91.4 & 92.5 & 94.2 & 86.2	 & 90.2	& 92.5 & 94.5 &	95.1 \\
		$PL(21, 25)$, $k=2$ & 81.1 & 87.0 & 91.8 & 93.2 & 94.5 & 87.2 & 90.0 & 92.9 & 94.3 & 95.0\\
		$PL(21, 25)$, $k=3$ & 80.1 & 86.3& 92.1 & 93.0& 94.2 & 86.6 & 90.3 & 92.8 & 94.2 & 94.7\\
		\hline
		$PL(66, 71)$, Crop & 78.4 & 87.0 & 92.1 & 92.5 & 93.5	 & 86.1 & 89.0 & 92.8 & 93.5	 & 95.0\\
		$PL(66, 71)$, $k=1$ & 80.1 & 87.0 & 92.5 & 93.5 & 94.9 & 86.7 & 90.3 & 92.9 & 94.6 & 95.0\\
		$PL(66, 71)$, $k=2$ & 80.5 & 87.7 & 92.5 & 93.8 & 95.2 & 87.2	 & 90.2 & 93.2 & 94.2 & 95.0\\
		$PL(66, 71)$, $k=3$ & \textbf{82.0} & \textbf{89.0} & \textbf{92.1} & \textbf{93.8} & \textbf{95.5} & \textbf{87.6} & \textbf{90.6} & \textbf{93.4} & \textbf{94.7} & \textbf{95.2} \\
		\hline
	\end{tabular}
\end{table*}

\begin{table*}[]
	\caption{Ablation of components in bipartite graph convolutional network on DDSM and in-house datasets (\%).}
	\label{ablation_relation}
	\centering
	\footnotesize
	\renewcommand{\arraystretch}{1.2}
	\begin{tabular}{ll|ccccc|ccccc}
		\hline
		\multirow{2}{*}{$H^g$} & \multirow{2}{*}{$H^s$} & \multicolumn{5}{c|}{\textbf{In-house}} & \multicolumn{5}{c}{\textbf{DDSM}} \\ \cline{3-12} 
		&  & \textbf{R@0.5} & \textbf{R@1.0} & \textbf{R@2.0} & \textbf{R@3.0} & \textbf{R@4.0} & \textbf{R@0.5} & \textbf{R@1.0} & \textbf{R@2.0} & \textbf{R@3.0} & \textbf{R@4.0} \\ \hline
		& & 76.7 & 83.9 & 89.4 & 91.4 & 91.8 & 84.0 & 88.3 & 91.7 & 93.2 & 94.5\\
		$\surd$ & &  77.7 & 86.3 & 89.4 & 91.8 & 93.5 & 84.7 & 88.9 & 92.1 & 93.4 & 94.0 \\
		& $\surd$ &  78.4 & 83.9 & 91.1 & 92.1 & 93.8 & 85.2 & 89.2 & 92.0 & 93.4 & 94.2  \\
		$\surd$ & $\surd$ & \textbf{79.5} & \textbf{86.6} & \textbf{91.8} & \textbf{92.5} & \textbf{94.5} & \textbf{86.2} & \textbf{89.5} & \textbf{92.5} & \textbf{93.6} & \textbf{94.6}\\
		\hline
	\end{tabular}
\end{table*}

\begin{table*}[]
	\caption{Ablation of components in inception graph convolutional network on  DDSM and In-house datasets (\%).}
	\label{ablation_ign}
	\centering
	\footnotesize
	\renewcommand{\arraystretch}{1.2}
	\begin{tabular}{c|ccccc|ccccc}
		\hline
		\multicolumn{1}{c|}{\multirow{2}{*}{\textbf{Method}}} & \multicolumn{5}{c|}{\textbf{DDSM}} & \multicolumn{5}{c}{\textbf{In-house}} \\ \cline{2-11}
		\multicolumn{1}{c|}{} & \multicolumn{1}{c}{\textbf{R@0.5}} & \multicolumn{1}{c}{\textbf{R@1.0}} & \multicolumn{1}{c}{\textbf{R@2.0}} & \multicolumn{1}{c}{\textbf{R@3.0}} &  \multicolumn{1}{c|}{\textbf{R@4.0}} & \multicolumn{1}{c}{\textbf{R@0.5}} & \multicolumn{1}{c}{\textbf{R@1.0}} & \multicolumn{1}{c}{\textbf{R@2.0}} & \multicolumn{1}{c}{\textbf{R@3.0}} & \textbf{R@4.0} \\ \hline
		IGN(1) &  78.0 & 84.5 & 90.4 & 92.1 & 93.8 & 84.5 & 89.5 & 92.4 & 94.1 & 94.9\\
		IGN(3) & 78.1 & 86.0 & 91.0 & 92.5 & 93.5 & 85.0 & 89.0 & 92.3 & 94.2 & 94.4 \\
		IGN(5) & 78.4 & 85.3 & 91.1 & 93.2 & 94.2 & 85.3 & 89.1 & 92.3 & 94.1 & 94.7 \\
		IGN(1, 3) & 78.8 & 85.8 & 91.1 & 93.2 & 94.2 & 85.2 & 89.3 & 92.4 & 94.2 & 94.9\\ 
		IGN(1, 3, 5) & \textbf{79.1} & \textbf{86.3} & \textbf{91.4} & \textbf{93.2} & \textbf{94.5} & \textbf{85.6} & \textbf{89.5} & \textbf{92.8} & \textbf{94.4} & \textbf{95.0} \\
		\hline
	\end{tabular}
\end{table*}

\begin{table*}[]
	\caption{Ablation of modules on DDSM and in-house datasets (\%).}
	\label{ablation_module}
	\centering
	\footnotesize
	\renewcommand{\arraystretch}{1.2}
	\begin{tabular}{ll|ccccc|ccccc}
		\hline
		\multirow{2}{*}{\textbf{BGN}} & \multirow{2}{*}{\textbf{IGN}} & \multicolumn{5}{c|}{\textbf{DDSM}} & \multicolumn{5}{c}{\textbf{In-house}} \\ \cline{3-12} 
		&  & \textbf{R@0.5} & \textbf{R@1.0} & \textbf{R@2.0} & \textbf{R@3.0} & \textbf{R@4.0} & \textbf{R@0.5} & \textbf{R@1.0} & \textbf{R@2.0} & \textbf{R@3.0} & \textbf{R@4.0} \\ \hline
		& & 84.0 & 88.3 & 91.7 & 93.2 & 94.5 & 76.7 & 83.9 & 89.4 & 91.4 & 91.8 \\
		$\surd$ &  & {86.2} & {89.5} & {92.5} & {93.6} & {94.6} & {79.5} & {86.6} & {91.8} & {92.5} & {94.5} \\
		& $\surd$ &  85.6 & 89.5	 & 92.8 & 94.4 & 95.0 & 79.1 & 86.3 & 91.4 & 93.2 & 94.5 \\
		$\surd$ & $\surd$ & \textbf{87.6} & \textbf{90.6} & \textbf{93.4} & \textbf{94.7}  & \textbf{95.2} & \textbf{82.0} & \textbf{89.0} & \textbf{92.1} & \textbf{93.8}  & \textbf{95.5} \\
		\hline
	\end{tabular}
\end{table*}

\begin{figure*}[t]
	\centering
	\subfigure[]{\label{fig:1}\includegraphics[height=0.84\linewidth, width=0.105\linewidth]{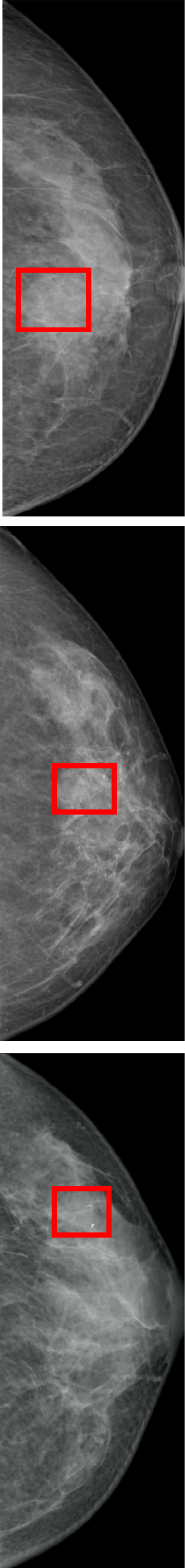}}
	\subfigure[]{\label{fig:2}\includegraphics[height=0.84\linewidth, width=0.105\linewidth]{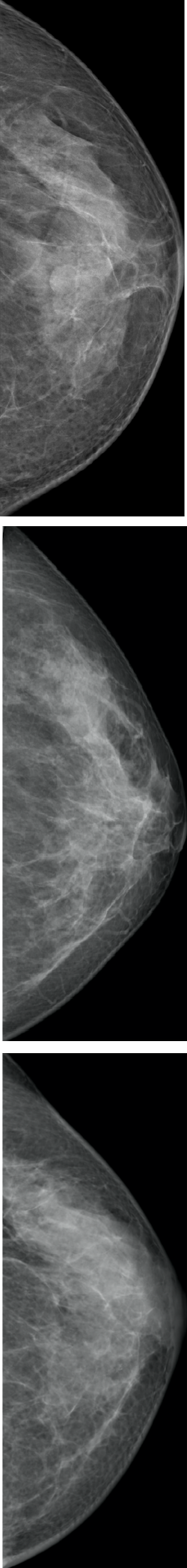}}
	\subfigure[]{\label{fig:6}\includegraphics[height=0.84\linewidth, width=0.105\linewidth]{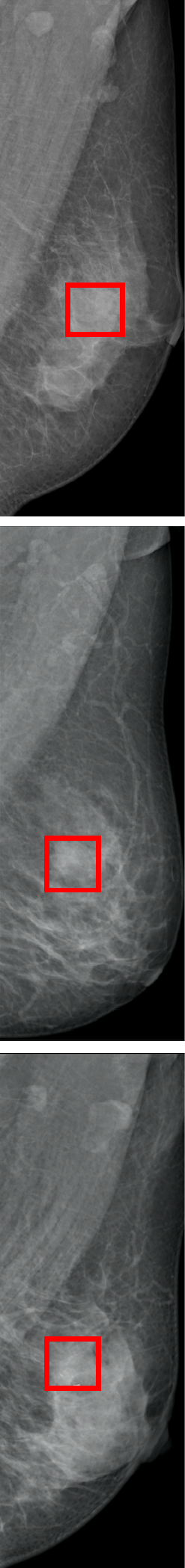}}
	\subfigure[]{\label{fig:7}\includegraphics[height=0.84\linewidth, width=0.105\linewidth]{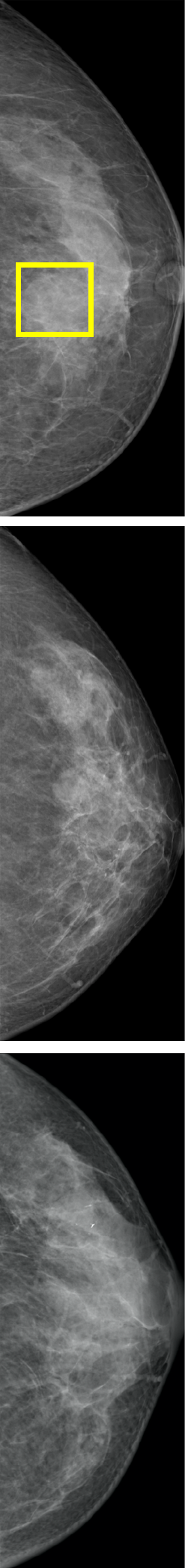}}
	\subfigure[]{\label{fig:3}\includegraphics[height=0.84\linewidth, width=0.105\linewidth]{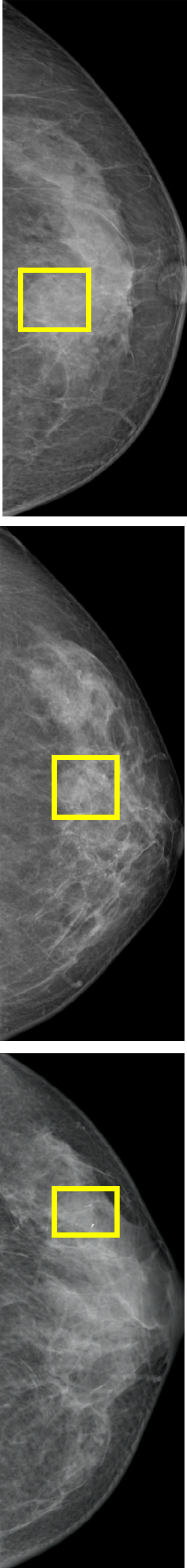}}
	\subfigure[]{\label{fig:4}\includegraphics[height=0.84\linewidth, width=0.105\linewidth]{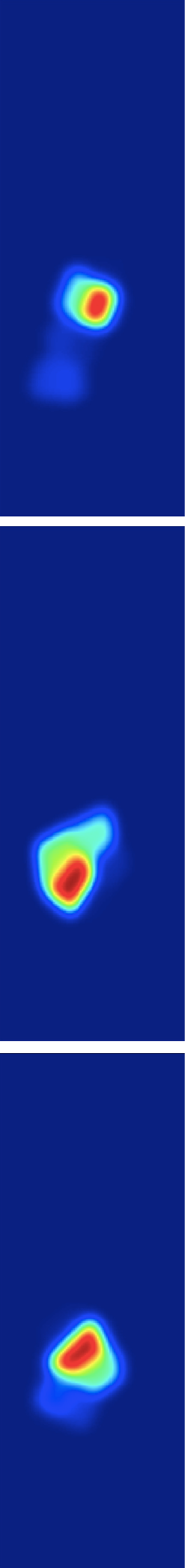}}
	\subfigure[]{\label{fig:5}\includegraphics[height=0.84\linewidth, width=0.105\linewidth]{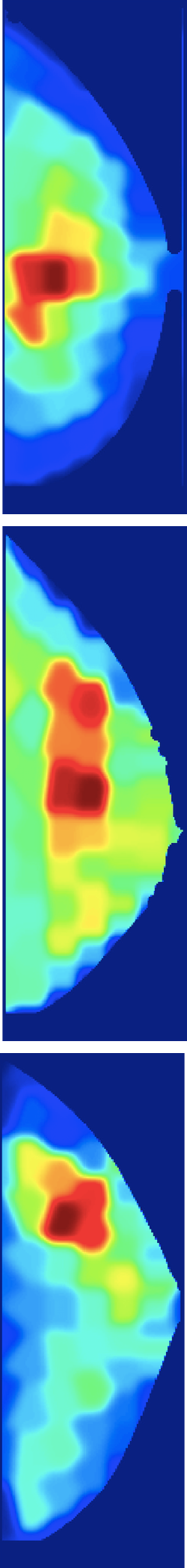}}
	\subfigure[]{\label{fig:5}\includegraphics[height=0.84\linewidth, width=0.105\linewidth]{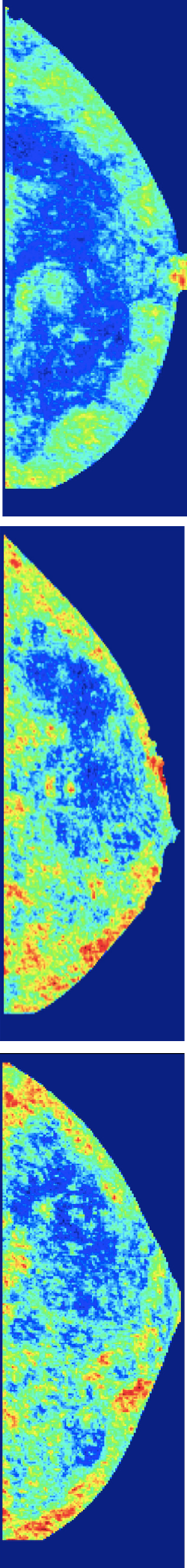}}
	\subfigure[]{\label{fig:5}\includegraphics[height=0.84\linewidth, width=0.105\linewidth]{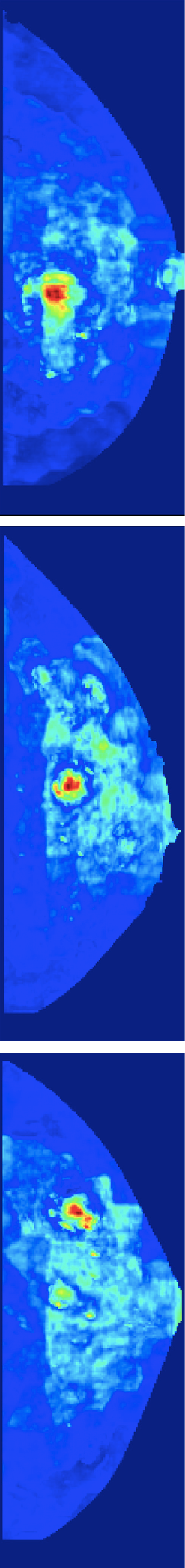}}
	
	\caption{Detection results of AG-RCNN. Each row shows a representative case. Column (a)-(c) refer to the examined view, the flipped contralateral view and the auxiliary view with annotations. Column (d)-(e) indicate detection results by Mask-RCNN and AG-RCNN. Column (f) visualizes the attention area on the auxiliary view. Column (g) shows the attention regions of bilateral views. Column (h)-(i) visualize the response maps before and after correspondence reasoning enhancement. }
	\label{fig:case_analysis}
\end{figure*}

\noindent \textbf{Mask RCNN, FPN, DCN. }  \ 
DCN is further integrated into the Mask RCNN baselines to improve the performance. 
\\

\noindent \textbf{CVR-RCNN.} \ 
CVR-RCNN~\cite{ma2019cross} models the ipsilateral relations of mammograms by adding a relation module~\cite{hu2018relation} into the second stage of detection process. 
\\

\subsection{Comparison with State-of-the-art Methods}
\label{comparison_with_sota}
We evaluate the performance by recall ($R$) at $t$ ($t \in \{0.5, 1.0, 2.0, 3.0, 4.0\}$) false positive per image (FPI), which is simplified as $R@t$. A mass region is recalled when its IOU (Intersection Over Union) value is larger than 0.2. 

Table~\ref{ddsm_report} and Table~\ref{ddsm_our} display the experimental results on DDSM dataset. Baseline results in Table \ref{ddsm_report} are cited from their original papers \cite{campanini2004novel, eltonsy2007concentric, sampat2008model, ma2019cross, Liu_2020_CVPR}. In Table~\ref{ddsm_our}, we re-implemented the baseline methods in our experiments. 
We do not make a comparison with \cite{liu2019unilateral} since their split of dataset is different. 
We keep the same FPI and compare with a strong baseline approach, CVR-RCNN~\cite{ma2019cross}. 
We can see that AG-RCNN significantly outperforms all compared methods.
The results on in-house dataset are reported in Table~\ref{in_house}.  
Although in-house dataset has larger amount of data and image modalities, our approach consistently outperforms all comparison methods, which verifies the effectiveness and robustness of AG-RCNN on the challenging scenario.

MommiNet~\cite{yang2020momminet} is an existing work on multi-view mammogram mass detection but with different experimental settings. 
Following the same practice of MommiNet, we randomly divide all cases on the DDSM dataset into the training, validation, and test sets by approximately 8:1:1, resulting in 8,256, 1,020 and 1,036 images in the respective sets. The proposed AG-RCNN outperform MommiNet by +1.5\% (@0.5), +1.8\% (@1.0), and +2.3\% (@2.0). 

To explore how the proposed model benefits from the correspondence reasoning mechanism, we qualitatively analyze some cases in Figure~\ref{fig:case_analysis}. 
As shown in the 2nd and 3rd rows, 
detecting mass lesions with only a single view is quite confusing since mass lesions are obscured by compacted glands in breasts.
Leveraging visual cues from different mammographic views can provide clear and reasonable evidences for mass detection in the examined view, and thus make the detection process more efficient and interpretable.
By doing so, the proposed method can significantly improve the recall. Besides, as shown in 1st row, the localization of the bounding box becomes more precise, since more detailed information about the anatomical structure of mass is taken into consideration.

\subsection{Ablation Study}
\label{ablation_study}


\subsubsection{Ablation of Pseudo Landmarks}
We compare the proposed pseudo landmarks with uniform grids, which embeds nodes uniformly into mammogram images without considering the visual prior knowledge. 
The number of nodes is identical between uniform grids and pseudo landmarks.  
The results are shown in Table~\ref{ablation_landmark}, which clearly demonstrate the superiority of our pseudo landmarks. 
We further investigate the effect of the number of nodes on the performance.  
``$PL(x, y)$'' in Table \ref{ablation_nb_node} denotes the setting that there are $x$ nodes in CC view and $y$ nodes in MLO view.
The setting ``$PL(1, 1)$'' is approximatively equivalent to two-branch Faster RCNN. 
The results are reported in Table \ref{ablation_nb_node}, we choose ``$PL(66, 71)$'' as our final results.

\subsubsection{Ablation of Graph Node Mapping.} 
To explore the effectiveness of bipartite node mapping, we first compare with a simple baseline, which directly crops a fixed region for graph node representation. 
Results in Table~\ref{ablation_bipartite_mapping} verify the superiority of the proposed graph node mapping. 
The justifications can be summarized as follows. First, the graph node mapping endows node representations with the property that the representative region of each node is invariant to the variations of image scales and breast shapes, which enhances the robustness of the node representation. Second, the graph node mapping mechanism is easy-to-implement without any additional parameters, which simplifies the training process.

We further analyze the relationship between $k$ and the results. 
We keep $k$ fixed for both $\phi_k$ and $\psi_k$. As shown in Table \ref{ablation_bipartite_mapping}, when the dense nodes are embedded, the model performs better with a larger $k$. The reason is that the dense nodes have smaller representative regions, and a larger $k$ can extract richer context features for each node.

\subsubsection{Ablation of Bipartite Graph Convolutional Network.} 
We conduct the ablation study of bipartite graph convolutional network based on ipsilateral views. 
To analyze the influence of each component in the bipartite graph convolutional network, we isolate each component (i.e., $H^g$ and $H^s$) of bipartite graph edges. 
We traverse all combinations of using $H^g$ and $H^s$.
It degenerates to a plain Mask RCNN when neither $H^s$ nor $H^g$ are used. The justification is that no information propagates across views and the prediction is only based on the examined view.
When either $H^s$ or $H^g$ is used, we set ${H}$ to ${H}^s$ or ${H}^g$ respectively, which propagates information across correspondences satisfying semantic constraints or geometric constraints only. 
The results are shown Table~\ref{ablation_relation}, which reveal two critical observations. 
First, compared with the single view based method, utilizing either ${H}^s$ or ${H}^g$ can provide considerable improvements. Second, by combining semantic and geometric relations, we can achieve the best performance.

\subsubsection{Ablation of Inception Graph Convolutional Network}
We conduct the ablation study of inception graph convolutional network based on bilateral views. For simplicity, ``IGN($s_1, s_2, s_3$)'' denote IGN with the settings of three different branches, i.e., $s_1, s_2, s_3$.
Table \ref{ablation_ign} reports the experimental results, which demonstrates the following conclusions. 
First, equipping the model with the tolerance of geometric distortions  will enhance the performance. Second, adopting multi-branch information propagation among top-s nearest nodes achieves the best performance.

\subsubsection{Ablation of Modularities}
We evaluate all combinations of BGN and IGN. 
The model degenerates to a plain Mask RCNN if neither BGN nor IGN are used, since the detection is only based on the examined view without multi-view information propagation. When using BGN only, the enhanced feature $Y$ is reformulated as:
\begin{equation}
	\begin{aligned}
		Y =[F_e, F_B] W_f^T.
	\end{aligned}
\end{equation}
While using IGN, the enhanced feature $Y$ with parameter $W_f \in \mathbb{R}^{C \times C}$ can be reformulated as:
\begin{equation}
	\begin{aligned}
		Y =(\hat{F}_I \cdot F_e) W_f^T.
	\end{aligned}
\end{equation}
The results are summarized in Table \ref{ablation_module}, which demonstrate that the performance gain benefits from both BGN and IGN.

\subsection{Visualization}
\label{visualization}
Our visualization experiments mainly answer three questions: (1) Where does the bipartite graph focus on auxiliary view? (2) Where does inception graph convolutional network focus on bilateral views? (3) How does the correspondence reasoning mechanism enhance the feature representations?

Firstly, we develop a specialized method for correspondence visualization to answer the first question. The major objective is to seek the representative regions of correlated nodes in the auxiliary view when given a query mass in the examined view. We define a one-hot representative vector $x \in \mathbb{R} ^ {|\mathcal{V}_{CC} \cup \mathcal{V}_{MLO} |}$ to denote the locations of the query masses in the examined area. The index of the node, which is nearest to the center of the analyzed mass in the examined view, is set to 1. We visualize the feature via Equation \ref{eq_vis}, where $o \in \mathbb{R} ^ {HW}$ stands for the response vector, and $[\cdot]_e$ represents indexing operator which selects nodes in the examined view from bipartite graph node set. 
We reshape and normalize the response vector $o$ to the output image. 
\begin{equation}
	\label{eq_vis}
	\begin{aligned}
		o = Q^r [H^B x]_e
	\end{aligned}
\end{equation}
As can be seen in Figure \ref{fig:case_analysis}~(f), we found that the bipartite graph focuses on the matched mass area in the auxiliary view, which is helpful for learning complementary feature representations. 
In addition, the proposed model has a clear physical meaning and provides visual cues of matched masses. Therefore, it is capable of assisting radiologists in clinical mammography interpretation. 

Secondly, we visualize the attention regions learned by the inception graph convolutional network. Since there exists a natural spatial attention map $\hat{F}_I$, we simply normalize it and obtain the visualization map. 
As illustrated in Figure \ref{fig:case_analysis} (g), attention regions mostly appear at asymmetric areas in bilateral views, which provides positive evidence of the regions to be mass lesions.

Lastly, to investigate how correspondence reasoning mechanism enhances the feature representations, we compare the response map before and after feature enhancement. To be specific, we respectively conduct channel-wise max pooling on $F_e$ and $Y$. 
The results are shown in Figure \ref{fig:case_analysis}. We can observe that feature response map activates more prominently on the mass region after enhancement. By doing so, the corresponding reasoning enhancement method helps to promote the detection performance and make a sufficient and comprehensive clinical decision.


\section{Conclusion}
\label{sec:conclusion}
In this paper, we delve into the multi-view correspondence reasoning problem and introduce a anatomy-aware graph convolutional network to endow the mammogram mass detection models with customized reasoning ability.
By jointly reasoning and distilling information from multiple mammography views, our model substantially enhances the expressive power of learned representations in the examined view during the detection process. 
The proposed model includes a bipartite graph convolutional network and an inception graph convolutional network. The former one is capable of performing reasoning about ipsilateral correspondences and modeling both geometric constraints and visual similarities across ipsilateral views, and the latter one can model the structural similarities between bilateral views.
To this end, correspondence reasoning enhancement propagates information through both graphs, which makes the spatial visual features aware of the multi-view correspondences.
Extensive experiments on both public and in-house datasets reveal that the proposed model significantly exceeds the state-of-the-art performance. In addition, visualization results show that AGN provides reasonable and interpretable visual cues for the clinical diagnosis. 


\section*{Acknowledgment}
This work was supported in part by Zhejiang Province Key Research \& Development Program (No. 2020C03073), MOST-2018AAA0102004, NSFC-61625201, 61527804, DFG TRR169 / NSFC Major International Collaboration Project "Crossmodal Learning". 

\bibliographystyle{IEEEtran}
\bibliography{IEEEabrv,ref}

\begin{IEEEbiography}[{\includegraphics[width=1in,height=1.25in,clip,keepaspectratio]{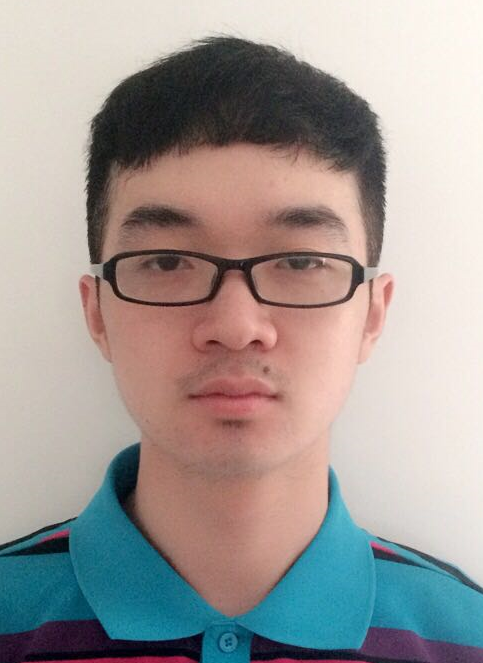}}]{Yuhang Liu}
	received the Bachelor degree in Computer Science and Technology from Xidian University in 2016, and the Master degree in Intelligence Science and Technology from Peking University in 2019. 
	He is a machine learning researcher at Deepwise AI Lab.
	His research interests include biomedical image analysis, deep learning and computer vision.
\end{IEEEbiography}

\begin{IEEEbiography}[{\includegraphics[width=1in,height=1.25in,clip,keepaspectratio]{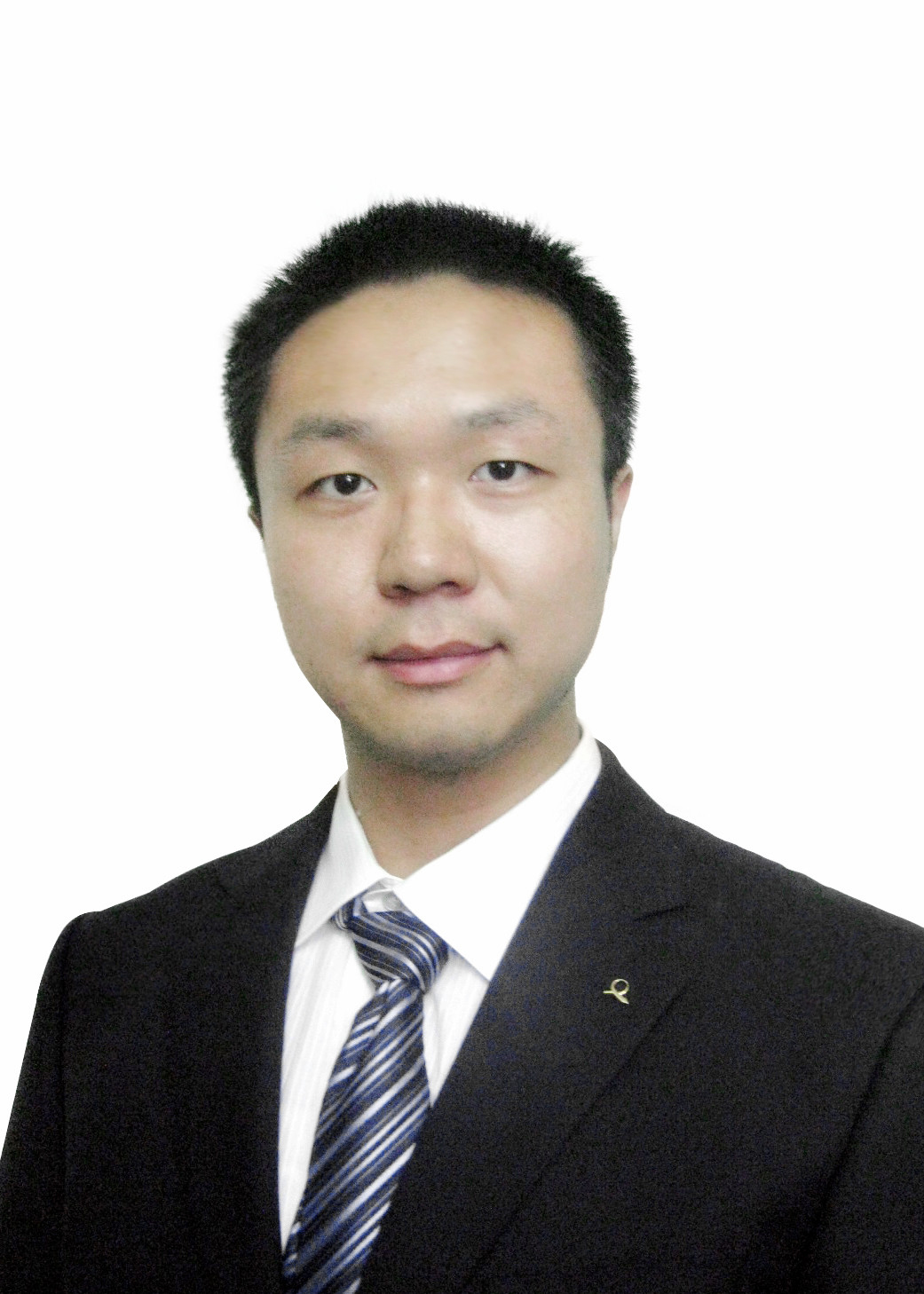}}]{Fandong Zhang}
	is currently a post-doc in Academy for Advanced Interdisciplinary Studies at Peking University. Previously, he received the Ph.D. degree in EECS from Peking University. His research interests include medical image analysis, computer vision and biometrics. Recently, he is working on automatic analysis of breast medical imaging, including mammography, MRI and Ultrasound.
\end{IEEEbiography}

\begin{IEEEbiography}[{\includegraphics[width=1in,height=1.25in,clip,keepaspectratio]{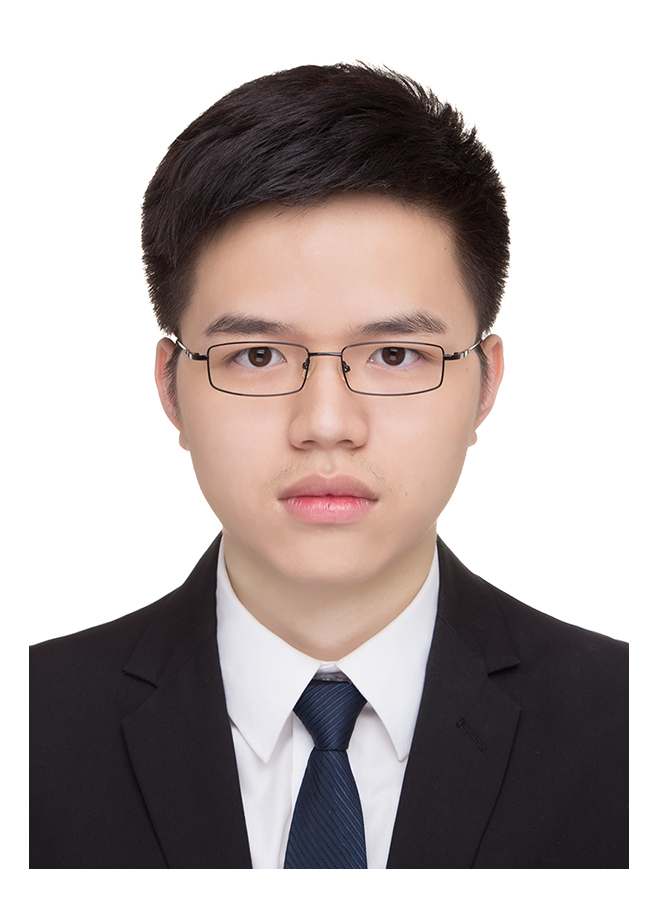}}]{Chaoqi Chen}
	received the B.S. and M.E. Degrees from Xiamen University, China, in 2017 and 2020, respectively. He is currently a Research Intern at Deepwise AI Lab, Beijing, China. His research spans computer vision and machine learning, with special interests in deep learning, transfer learning, and their visual applications.
\end{IEEEbiography}

\begin{IEEEbiography}[{\includegraphics[width=1in,height=1.25in,clip,keepaspectratio]{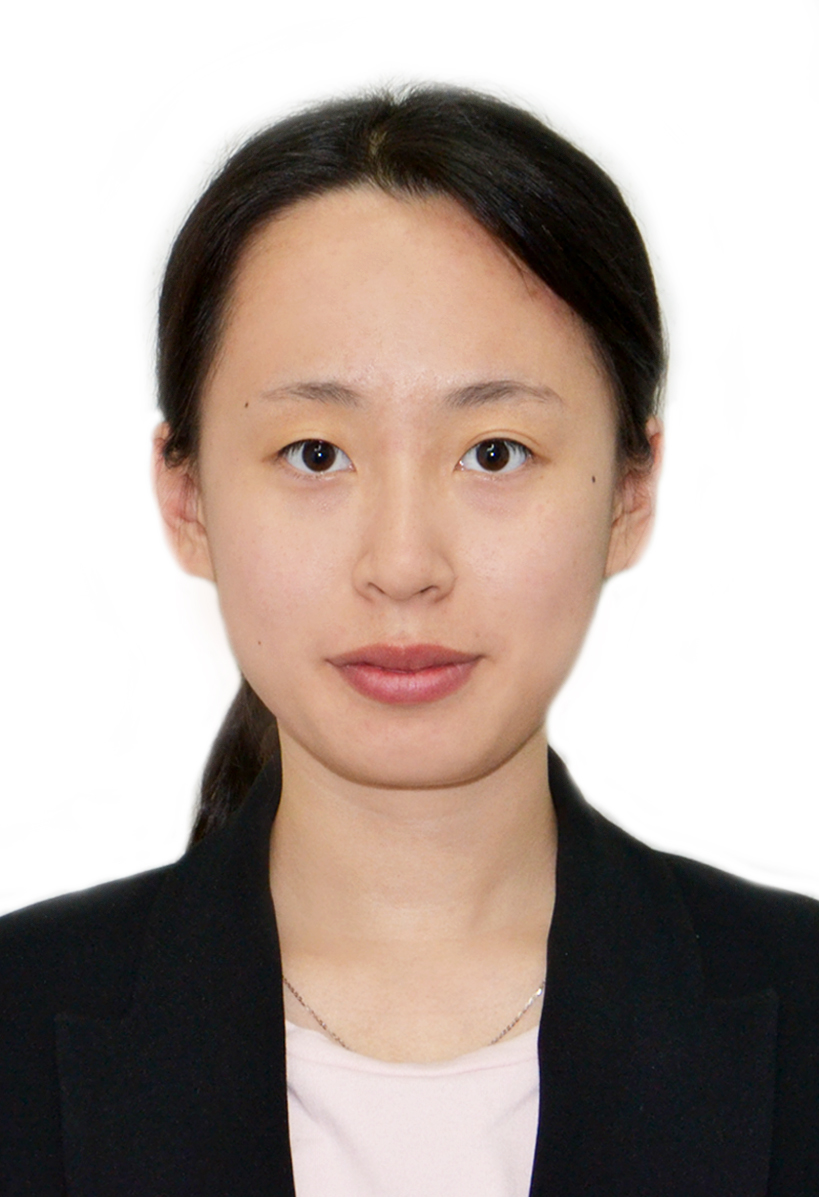}}]{Siwen Wang}
	received the BS and MS degrees in information and communication engineering from Dalian University of Technology, Dalian, China, in 2017 and 2019, respectively. She is currently with the Deepwise AI Lab, Beijing, China. Her current research interests include computer vision, pattern recognition, and medical image analysis.
\end{IEEEbiography}

\begin{IEEEbiography}[{\includegraphics[width=1in,height=1.25in,clip,keepaspectratio]{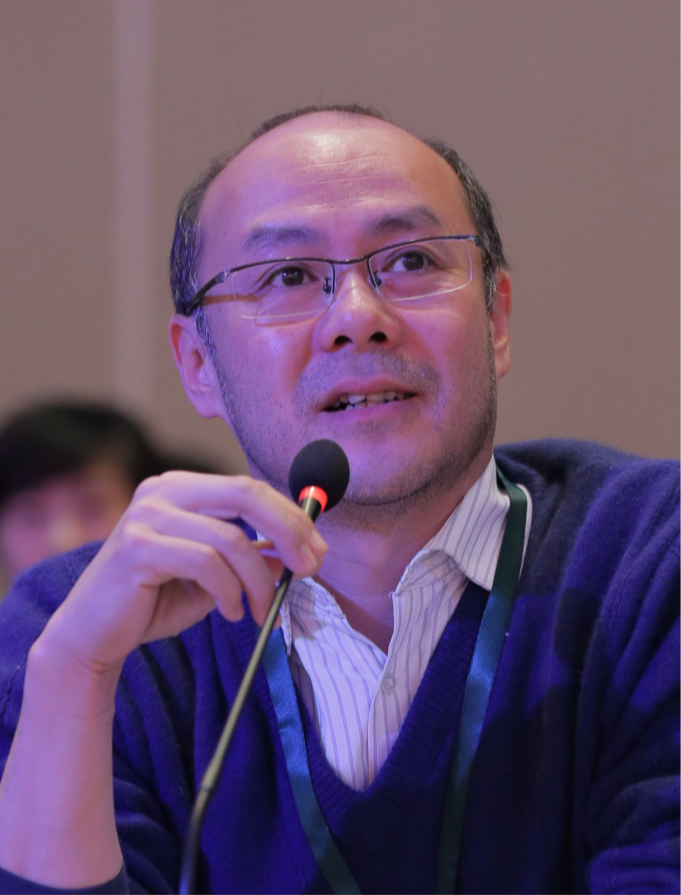}}]{Yizhou Wang}
	is a BoYa Professor of Computer Science Department and the vice director of the Center on Frontiers of Computing Studies at Peking University. He received his Bachelors degree in Electrical Engineering from Tsinghua University in 1996, and his Ph.D. in Computer Science from University of California at Los Angeles (UCLA) in 2005. He joined Xerox Palo Alto Research Center (Xerox PARC) as a research staff from 2005 to 2007. He was granted the National Natural Science Fund (NSFC) for Distinguished Young Scholars. Dr. Wang’s research interests include computational vision, statistical modeling and learning, medical image analysis, and digital visual arts.
\end{IEEEbiography}

\begin{IEEEbiography}[{\includegraphics[width=1in,height=1.25in,clip,keepaspectratio]{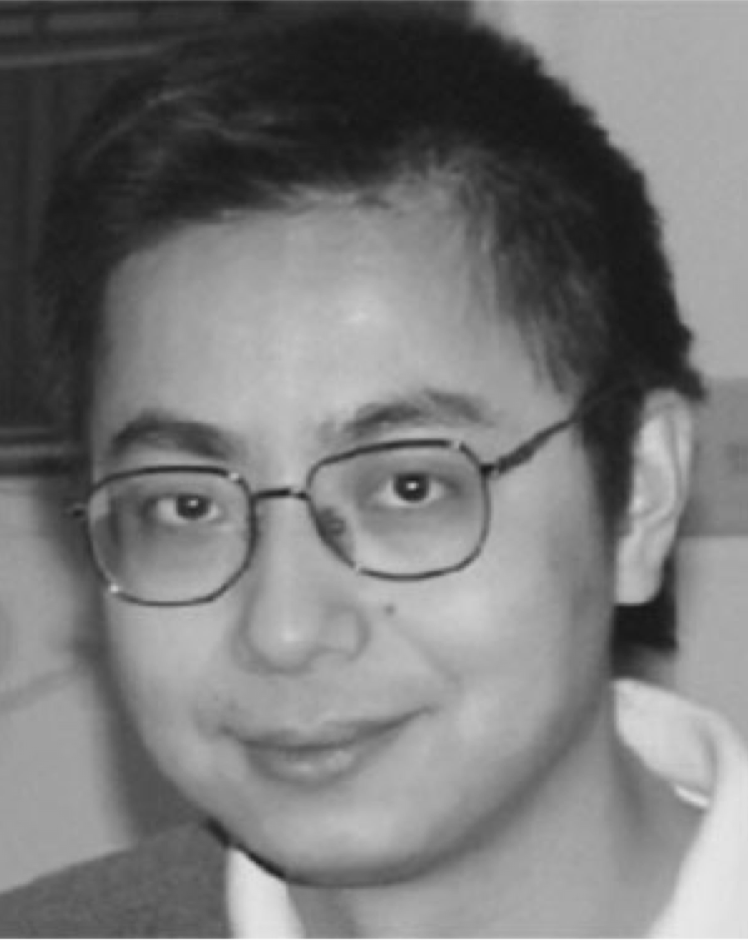}}]{Yizhou Yu}
	(M’10, SM’12, F’19) received the PhD degree from University of California at Berkeley in 2000. He is a professor at the University of Hong Kong, and also the chief scientist at Deepwise Healthcare. He was a faculty member at University of Illinois at Urbana-Champaign for twelve years. He is a recipient of 2002 US National Science Foundation CAREER Award and ACCV 2018 Best Application Paper Award. Prof Yu has served on the editorial board of IET Computer Vision, The Visual Computer, and IEEE Transactions on Visualization and Computer Graphics. He has also served on the program committee of many leading international conferences, including CVPR, ICCV, and SIGGRAPH. His current research interests include computer vision, deep learning, biomedical data analysis, computational visual media and geometric computing.
\end{IEEEbiography}

\end{document}